\documentclass[twocolumn]{article}

\usepackage{amsmath}
\usepackage{amsfonts}
\usepackage{amssymb}
\usepackage{microtype}
\usepackage{graphicx}
\usepackage{subcaption}
\usepackage{booktabs} % for professional tables
\usepackage{xcolor}
\usepackage{multirow}
\usepackage{float}

\usepackage{hyperref}

\usepackage{mathtools}
\usepackage{amsthm}

\definecolor{blue}{RGB}{139,167,206}
\definecolor{green}{RGB}{155,194,134}
\definecolor{purple}{RGB}{171,144,185}

\usepackage[capitalize,noabbrev]{cleveref}

\theoremstyle{plain}

\theoremstyle{definition}

\theoremstyle{remark}

\usepackage[textsize=tiny]{todonotes}

\title{TabSurv: Adapting Modern Tabular Neural Networks to Survival Analysis}

\author{
Stanislav Kirpichenko$^{*}$ \quad
Andrei Konstantinov$^{*,\dagger}$ \quad
Lev Utkin$^{*}$\\
\href{mailto:lkirpichenko.sr@gmail.com}{kirpichenko.sr@gmail.com},
\href{mailto:andrue.konst@gmail.com}{andrue.konst@gmail.com},
\href{mailto:lev.utkin@gmail.com}{lev.utkin@gmail.com}\\%[0.5em]
\small Higher School of Artificial Intelligence Technologies,\\
Peter the Great St. Petersburg Polytechnic University,\\
St. Petersburg, Russia\\
}

\date{}

\begin{document}

\maketitle

\begin{abstract}
  Survival analysis on tabular data is a well-studied problem. However, existing deep learning methods are often highly task-specific, which can limit the transfer of new approaches from other domains and introduce constraints that may affect performance. We propose TabSurv, an approach that adapts modern tabular architectures to survival analysis using either the Weibull distribution or non-parametric survival prediction. TabSurv optimizes SurvHL, a novel histogram loss function supporting censored data. In addition to a baseline feed-forward network, we implement deep ensembles of MLPs for survival analysis within TabSurv. In contrast to prior work, the ensemble components are trained in parallel, optimizing survival distribution parameters before averaging, which promotes diversity across ensemble component predictions. We perform a comprehensive empirical evaluation of different proposed architectures on 10 diverse real-world survival datasets. Our results show that TabSurv consistently outperforms on average established classical and deep learning baselines, such as RSF, DeepSurv, DeepHit, SurvTRACE. Notably, deep ensembles with Weibull parametrization instead of non-parametric models achieve the highest average rank by C-index. Overall, our study clarifies how modern tabular neural networks can be adapted and trained to tackle survival analysis problems, offering a strong and reliable approach. The TabSurv implementation is publicly available.
\end{abstract}

\section{Introduction}

Survival analysis \cite{Hosmer-Lemeshow-May-2008}, the statistical study of
time-to-event data, is a foundational tool for domains where outcomes are not
immediate and observations can be incomplete. Its ability to formally
incorporate censored data, where the exact event time for some instances may
only be known to exceed a certain point, makes it essential for clinical
trials, reliability engineering, financial risk assessment, and beyond.
Survival analysis delivers quantifiable insights into event timing uncertainty
across these fields.

While classical methods like the Kaplan-Meier estimator \cite{Kaplan-Meier-58}
and Cox Proportional Hazards models \cite{Cox-1972} and Random Survival
Forests (RSF) \cite{Ishwaran-Kogalur-2007} have dominated the field for
decades, recent years have seen significant interest in deep learning
approaches, including DeepSurv \cite{Katzman-etal-2018}, DeepHit
\cite{Lee-Zame-etal-2018}, and SurvTRACE \cite{wang2022survtrace}, which aim
to capture complex, non-linear relationships in high-dimensional data. In
parallel, nonparametric kernel-based methods, such as the Beran estimator
\cite{Beran-81}, provide flexible alternatives by smoothing the empirical
survival distribution based on covariate similarity, offering robustness to
model misspecification.

Despite this progress, a notable gap persists between advancements in generic
tabular deep learning, including TabR \cite{gorishniy2024tabr}, TabM
\cite{Gorishniy2024TabMAT}, RealMLP \cite{HolzmllerRealMLPAM}, etc., and their
application to survival analysis. Many existing deep survival models are
highly task-specific, with architectures and loss functions deeply intertwined
with the survival objective \cite{kvamme2021continuous}. This specialization
can inadvertently limit the field: it restricts the transfer of modern
architectures from the broader tabular machine learning domain and may impose
modeling constraints that affect flexibility and performance. Some approaches,
such as DeepSurv, inherit the proportional hazards assumption from the Cox
model, which may not hold in complex real-world scenarios. Others, like
DeepHit, relax this assumption but rely on discretizing the time axis, which
introduces sensitivity to binning choices and can impair calibration. More
recent models, including SurvTRACE, incorporate transformer-based
architectures but are often computationally intensive and require large
datasets to realize their full potential, conditions rarely met in many
survival analysis applications, which typically involve modest sample sizes
and high-dimensional covariates. Moreover, these models often lack explicit modular design, making it difficult to swap components (e.g.,
backbone networks, loss functions, or output parameterizations) without
significant re-engineering.

To address this, we propose \emph{TabSurv}, a principled and flexible approach
that bridges modern tabular deep learning with survival analysis. Our core
contribution is a framework that adapts standard and ensemble-based tabular
architectures, such as Multi-Layer Perceptrons (MLPs), to model survival
outcomes through two distinct pathways: a Weibull-distribution-based
parametric head and a non-parametric survival prediction head. A key
peculiarity of TabSurv is the introduction and optimization of a novel
histogram-based loss function, called SurvHL, explicitly designed to handle
censored data efficiently. Furthermore, we implement a parallel deep ensemble
strategy within TabSurv, where each ensemble member independently learns
survival distribution parameters before aggregation. Unlike sequential or
bagging methods, our ensemble components optimize their respective survival
distribution parameters independently before aggregation, ensuring prediction
diversity and enhanced robustness.

We conduct a comprehensive empirical evaluation of TabSurv variants on several
diverse real-world datasets. Our results demonstrate that TabSurv consistently
outperforms established classical (e.g., RSF) and deep learning baselines
(e.g., DeepSurv, DeepHit, SurvTRACE). The deep ensemble variant with Weibull
parametrization achieves the highest average rank by the concordance index
(C-index \cite{Harrell-15}), highlighting the efficacy of combining modern
ensemble techniques with a flexible parametric model. By providing a modular,
well-engineered, and publicly available implementation, TabSurv not only
advances the state of the art in deep survival analysis, but also offers a
reliable foundation for future research and practical deployment. Our work
clarifies how off-the-shelf tabular architectures can be systematically
adapted to handle censored time-to-event data, opening new avenues for
cross-domain knowledge transfer in survival modeling.

Our contributions can be summarized as follows:

\begin{enumerate}
\item We propose TabSurv, a flexible approach that adapts modern tabular deep
learning architectures to survival analysis, supporting both parametric
(Weibull) and non-parametric survival prediction while naturally handling
censored data.

\item We propose SurvHL, a novel histogram-based loss function tailored for
censored time-to-event data, and develop a parallel training strategy for deep
ensembles of MLPs that optimizes survival distribution parameters before
averaging, enhancing diversity and predictive performance.

\item Through extensive experiments on diverse real-world survival datasets,
we demonstrate that TabSurv consistently outperforms established classical and
deep learning baselines, including RSF, DeepSurv, DeepHit, SurvTRACE, with
Weibull-based deep ensembles achieving the highest average ranks of survival
performance measures, providing new insights into the effectiveness of
parametric modeling in deep survival analysis. To summarize the experimental
results, the C-index \cite{Harrell-15}, Integrated Brier Score (IBS)
\cite{graf1999assessment}, and time-dependent (cumulative/dynamic) AUC
\cite{Uno-etal-11} metrics are applied.
\end{enumerate}

In summary, this study clarifies how contemporary tabular neural networks can
be effectively adapted for survival analysis, providing a strong, reliable,
and publicly available approach. The complete implementation of TabSurv is
open-sourced to facilitate further
research and application.

\section{Related Work}

\textbf{Machine learning for survival analysis.} Survival analysis concerned
with modeling time-to-event outcomes under censoring has benefited
substantially from advances in machine learning. Early efforts extended
classical statistical models (e.g., Cox regression) with nonparametric and
ensemble methods, while recent work increasingly leverages deep learning to
capture complex, nonlinear relationships in high-dimensional data.
Comprehensive surveys by
\cite{marinos2021survey,salerno2023high,Wang-Li-Reddy-2019} catalog this
evolution, highlighting trends such as scalable deep architectures, handling
of competing risks, and integration with electronic health records.
Methodological generalizations of survival models are discussed in
\cite{bender2020general}, while \cite{EmmertStreib-Dehmer-19} emphasize
benchmarking and interpretability. Most recently, \cite{Wiegrebe:2024aa}
outline emerging challenges and opportunities in the field, including
robustness, fairness, and uncertainty quantification. 

Transformer-based models capable of modeling long-range
interactions in high-dimensional inputs have gained traction in survival
analysis: \cite{Wang-Sun-22} propose transformers for competing-risk settings,
while others develop specialized variants such as
ResDeepSurv~\cite{wang2024resdeepsurv}, explainable
transformers~\cite{tang2023explainable}, and general-purpose survival
transformers~\cite{hu2021transformer}. Applications in clinical domains
include multi-modal survival prediction~\cite{yao2024multi}, semi-supervised
learning~\cite{teng2025semi}, and SurFormer~\cite{wang2023surformer}. 
The discussed transformer survival models, in contrast to the proposed approach TabSurv, 
are based on learning from images rather than from tabular data.

\textbf{Modern tabular machine learning models}. A distinctive feature of many
machine learning tasks is that the data are predominantly tabular, comprising
lab results, patient characteristics, and similar structured variables. This
necessitates the use of models specifically designed for tabular data.
Prominent examples include TabR~\cite{gorishniy2024tabr}, TabM
\cite{Gorishniy2024TabMAT}, RealMLP \cite{HolzmllerRealMLPAM},
TabICL~\cite{qu2025tabicl}, TabPFN~\cite{hollmann2022tabpfn},
MotherNet~\cite{mueller2024mothernet}, along with several other recent
architectures~\cite{hollmann2025accurate,kim2025table,ma2025foundation,rubachev2025finetuning,xu2025mixture}. 
Ensembles of tabular models, where multiple neural networks are trained individually and their predictions are averaged, were studied in \cite{laurent2023packed,Wen2020BatchEnsemble}. 
Despite the effectiveness of these models with tabular data, their
application in survival analysis requires the development of special
``extensions'' that would enable the creation of new survival models competitive
with existing deep learning survival models.

\section{Survival analysis: the problem statement}

In survival analysis, a dataset is represented by a set of triplets
$\mathcal{D}=\{(\mathbf{x}_{1},\delta_{1},t_{1}),\dots,(\mathbf{x}_{n}%
,\delta_{n},t_{n})\}$, where $\mathbf{x}_{i}\in \mathbb{R}^{d}$ is a
$d$-dimensional feature vector, $t_{i}$ is the observed time, and $\delta
_{i}\in \{0,1\}$ is a binary censoring indicator. An observation is
\emph{uncensored} ($\delta_{i}=1$) if the event of interest (e.g., failure or
death) was observed at time $t_{i}$. It is \emph{censored} ($\delta_{i}=0$) if
the event was not observed within the study period, meaning the true event
time is known only to exceed $t_{i}$. A survival model is trained on
$\mathcal{D}$ to estimate probabilistic characteristics of the future event
time $T$ for a new subject described by a feature vector $\mathbf{x}$.

A fundamental concept in survival analysis is the \emph{survival function}
$S(t\mid \mathbf{x})$, defined as the probability that a subject with features
$\mathbf{x}$ survives beyond time $t$:
\begin{equation}
S(t\mid \mathbf{x})=\mathbb{P}(T > t \mid \mathbf{x}).
\end{equation}

Our objective is to estimate the survival function $S(t \mid x)$.
To this end, we discretize time into a set of intervals and represent the survival function via interval probabilities $p_i(x)$, defined as
\[
p_i(x) = \mathbb{P}\bigl(T \in [\tau_{i}, \tau_{i+1}) \mid X = x\bigr),
\]
where $0 < \tau_1 < \dots < \tau_m < \tau_{m + 1} = +\infty$ denote the ordered unique event times.
Given $\{p_i(x)\}_{i=1}^m$, the survival function can be approximated as
\[
S(t \mid x) \approx 1 - \sum_{i: \tau_i \le t} p_i(x).
\]
We denote the interval probabilities, predicted by a machine learning model as $\widehat{p}_i(\mathbf{x})$, and represent the estimated survival function as a vector $\widehat{\mathbf{S}}=(\widehat{S}_1, \dots, \widehat{S}_m)$, whose entries $\widehat{S}_i(\mathbf{x}) = \sum_{j=1}^i \widehat{p}_j(x)$ approximate $S(\tau_i \mid \mathbf{x})$.

\section{TabSurv}

\begin{figure*}[t]
    \centering
    \includegraphics[width=\textwidth]{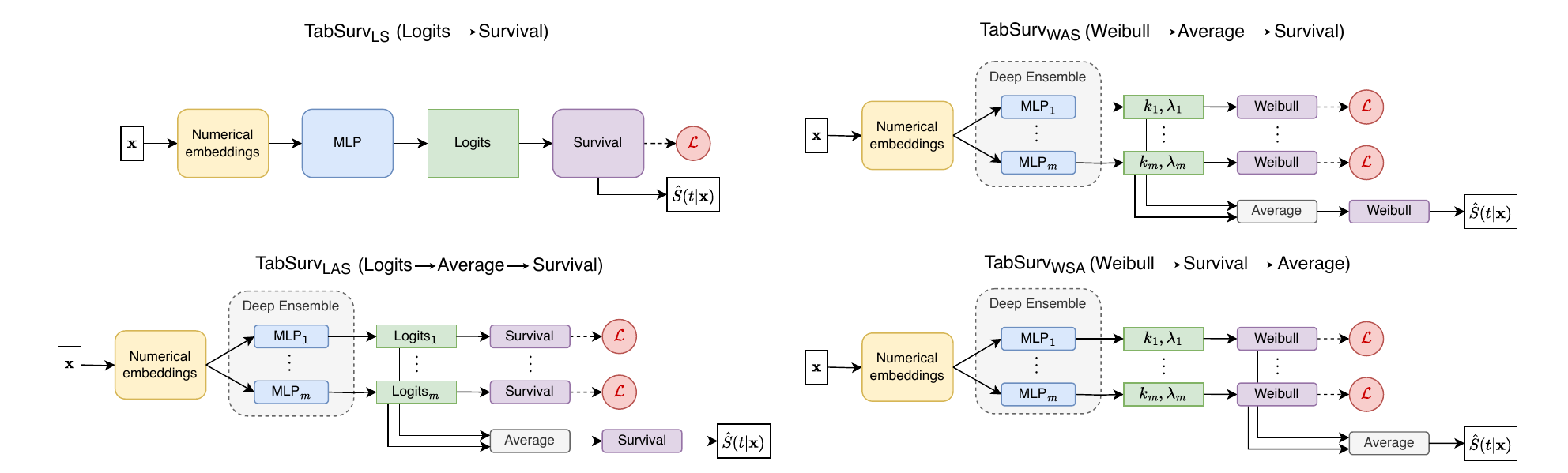}
        \caption{Architectural variants of TabSurv. \textbf{\color{blue}Blue} blocks denote learnable neural networks, \textbf{\color{green}Green} blocks denote intermediate parameters, and \textbf{\color{purple}Purple} blocks denote the specific survival transformation (Softmax or Weibull). \textbf{Dashed} arrows indicate the flow during training (independent losses), while \textbf{Solid} arrows indicate the prediction flow (including ensemble averaging).}
    \label{fig:fullwidth}
\end{figure*}

%\paragraph{Overview.}
In general, TabSurv consists of three main components: a numerical feature embedding module, a neural network backbone, and a survival function construction module.
Figure~\ref{fig:fullwidth} illustrates four proposed architectural variants, starting from the simplest MLP wrapper to deep ensemble, generating Weibull distribution parameters.
Below, we describe these variants in detail.

\subsection{Architecture details}

\paragraph{Logits $\rightarrow$ Survival (LS).}
The simplest approach is non-parametric, i.e., it does not assume any predefined family of event-time distributions.
A neural network of arbitrary architecture, such as a fully connected network or RealMLP, outputs logits corresponding to a discrete set of time intervals.
These logits are transformed into a probability distribution over time intervals via the Softmax function, which is then converted into a survival function.

\paragraph{Logits $\rightarrow$ Average $\rightarrow$ Survival (LAS).}
Unlike classical neural-network-based approaches to survival analysis, we additionally consider deep ensembles, where the neural network consists of $m$ independently trained components.
As shown in \cite{Gorishniy2024TabMAT}, such ensembles can lead to substantially more accurate models.
Technically, training is performed in parallel by minimizing the sum of the loss functions of all ensemble members.
Importantly, prediction aggregation is not performed before loss computation, which is crucial for preserving prediction diversity.
Indeed, optimizing the sum of losses over independent components is equivalent up to the shared modules to training $m$ ensemble members independently.

A key difference from classical ensembles is the presence of a shared numerical feature embedding module, which is trained jointly for all ensemble members.
This design choice is motivated by the need to balance ensemble diversity with representation efficiency.
Low-level numerical feature representations are largely task-agnostic and benefit from being learned from the full data signal, whereas higher-level predictions remain independent across ensemble members.
Sharing the embedding module reduces the number of trainable parameters, improves sample efficiency, and stabilizes optimization, while still allowing ensemble members to maintain diversity through independent downstream networks.

In the LAS architecture, each ensemble member outputs logits.
During training, these logits are independently transformed into survival functions, and a separate loss is computed for each member.
At inference time, the logits are averaged and subsequently converted into a survival function.
Thus, the final model represents an ensemble, which can potentially improve generalization performance.

\paragraph{Weibull parameters $\rightarrow$ Survival $\rightarrow$ Average (WSA).}
Since non-parametric approaches may yield noisy survival function estimates, we also consider a parametric formulation based on the Weibull distribution.
We adopt the Weibull distribution as a parametric survival model due to its favorable balance between flexibility and inductive bias.
The Weibull family can represent increasing, decreasing, and constant hazard rates, which makes it expressive enough for a wide range of real-world time-to-event processes while remaining low-dimensional and stable to optimize.
Compared to more expressive parametric families, the Weibull distribution exhibits well-behaved likelihood surfaces and requires estimating only two positive parameters, $\lambda$ and $k$.
This significantly reduces estimation variance and improves robustness, especially in low- and moderate-sample regimes.
Finally, the Weibull distribution admits a closed-form cumulative distribution and survival function, which allows for efficient discretization on arbitrary time grids and seamless integration with standard survival losses.
Although this parameterization is not universal, in practice one can determine its suitability for a given task via cross-validation.
As demonstrated in our experiments, the Weibull-based approach achieves superior performance on a wide range of real-world datasets.

In the WSA architecture, each ensemble member predicts positive Weibull parameters $\lambda(x)$ and $k(x)$.
Based on these parameters, we compute a discretized probability distribution, where each element represents the probability that the event occurs inside the interval $[\tau_{i}, \tau_{i + 1})$:
\begin{equation}
\widehat{p}_i(x) = F_W(\tau_{i+1}; \lambda(x), k(x)) - F_W(\tau_{i}; \lambda(x), k(x)),
\end{equation}
\begin{equation}
\widehat{S}_i(x) = 1 - \sum_{j=1}^{i} \widehat{p}_j(x),
\end{equation}
where $F_W(\cdot)$ denotes the Weibull cumulative distribution function.

Training proceeds analogously to LAS: each ensemble member is optimized using its own loss function.
At inference time, the estimated survival functions are averaged.
Consequently, the final survival estimate is a mixture of the survival functions produced by individual ensemble members.

\paragraph{Weibull parameters $\rightarrow$ Average $\rightarrow$ Survival (WAS).}
The WAS architecture differs from WSA only at the inference stage; however, this difference is substantial.
Averaging the Weibull parameters prior to survival function computation results in a final distribution that remains Weibull.
While this imposes a strong modeling constraint, it can significantly reduce prediction noise.

Empirically we observe that the multimodality produced by WSA is more likely to reflect estimation noise rather than meaningful uncertainty.
Since each MLP is trained independently by fitting the whole distribution, aggregated multimodal survival time estimates cannot represent a complex multimodal distribution.
Therefore, WSA should not be regarded as inherently superior to WAS.

%\paragraph{Important practical modifications.}

\subsection{SurvHL loss function}

The idea behind the loss function SurvHL is to extend the discretized likelihood commonly used in survival analysis,
which is defined by probabilities on discretized time intervals in
accordance with a conditional probability distribution of time to event and
represented in survival analysis or a single data point $\mathbf{x}_{i}$ as
follows:
\begin{equation}
%\ell(\mathbf{x}_{i},T_{i},\delta_{i})=\delta_{i}\sum \limits_{j=1}^{M}\log
%f(t_{j}\mid \mathbf{x}_{i})+(1-\delta_{i})\log S(T_{i}\mid \mathbf{x}%
%_{i}).\label{loss_single_0}%
\ell(\mathbf{x}, t, \delta) = \begin{cases}
- \log \widehat{p}_{I(t)}(\mathbf{x}), & \delta = 1 \\
- \log \widehat{S}_{I(t)}(\mathbf{x}), & \delta = 0,
\end{cases}
\end{equation}
where $I(t) = \max \{ i \mid \tau_i \le t \}$ is the index of the corresponding time interval.

For uncensored events, instead of considering only one discretized time
interval (as in classical likelihood), we smooth across several neighboring
intervals. This approach ensures a smoother gradient flow through the loss
function and makes model training more robust compared to the standard maximum
likelihood method.

The formula for a single data point $\mathbf{x}$, for which the predicted probabilities and the survival function vector are
denoted for brevity as $\widehat{p}$, $\widehat{S}$:
\begin{equation}
%\ell(\mathbf{x}_{i},T_{i},\delta_{i})=\delta_{i}\sum \limits_{j=1}^{M}%
%\omega(t_{j},T_{i},\sigma^{2})\cdot \log f(t_{j}|X=\mathbf{x}_{i}%
%)+(1-\delta_{i})\log S(T_{i}|X=\mathbf{x}_{i}),\label{loss_single}%
\ell(\mathbf{x}, t, \delta) = \begin{cases}
-\sum_{j=1}^{m} \omega_j(I(t)) \log(\widehat{p}_{j}), & \delta = 1 \\
-\log \widehat{S}_{I(t)}, & \delta = 0,
\end{cases}
\end{equation}
where $\omega_j(I(t))$ are weights derived from the Gaussian
distribution centered at $I(t)$ with the standard deviation $\sigma = r/3$, and $r$ is a hyperparameter.
Gaussian weighting is chosen for smoothness and locality. %, we found it more stable than uniform or triangular kernels.

The final loss function is the negative average over a batch of size $n$:
\begin{equation}
\label{loss_total}\mathcal{L}=-\frac{1}{n}\sum \limits_{i=1}^{n}\ell
(\mathbf{x}_i, t_i, \delta_i).
\end{equation}

\begin{figure}[ptb]
\centering
\includegraphics[width=0.5\linewidth]{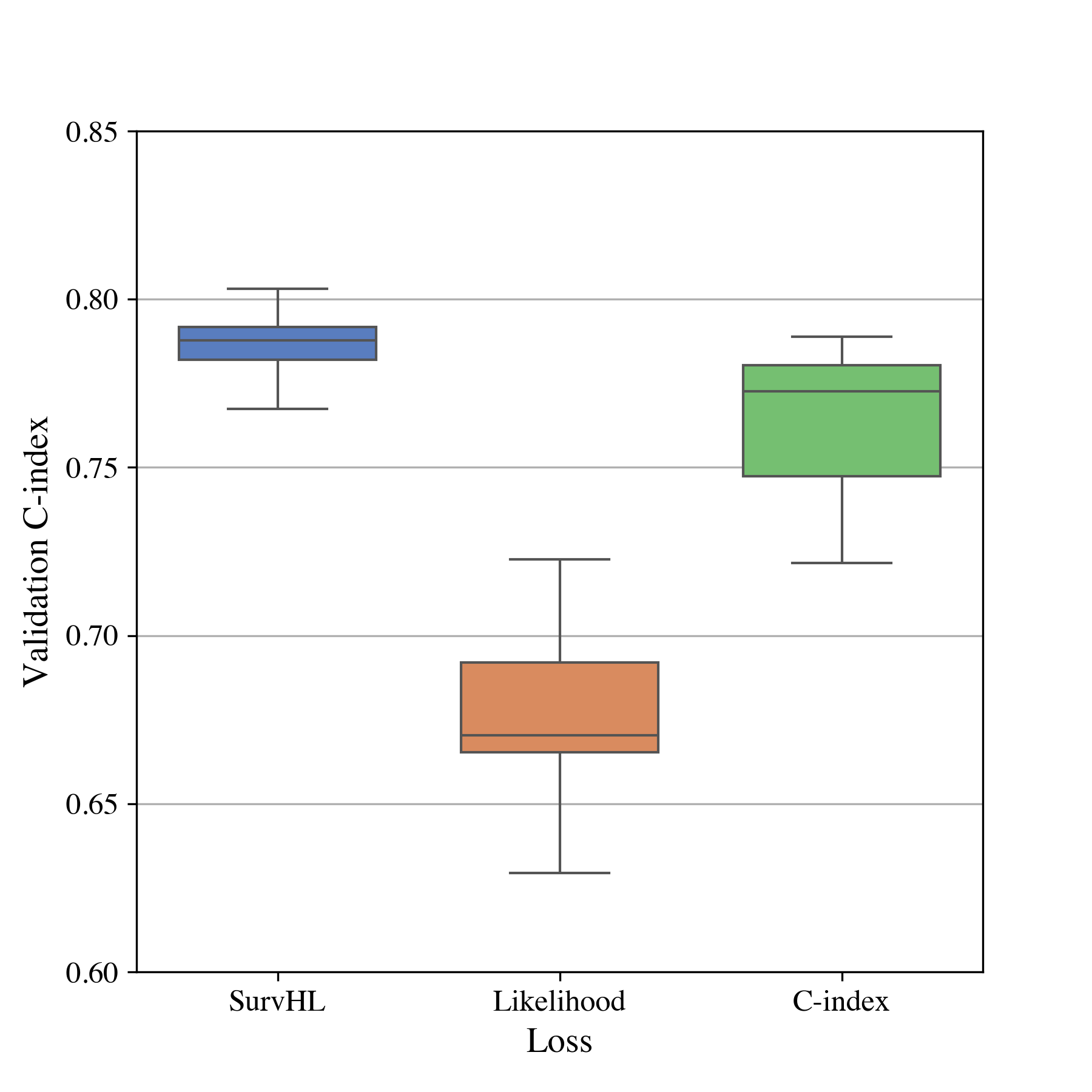}\caption{Hyperparameter
tuning results with different loss functions, TabSurv\textsubscript{LAS} model on METABRIC dataset}%
\label{fig:loss_boxplot}%
\end{figure}

Figure~\ref{fig:loss_boxplot} presents the distribution of the C-index on the validation set obtained during hyperparameter tuning of the TabSurv\textsubscript{LAS} model on the METABRIC dataset.
Among the considered objectives, the SurvHL consistently achieves higher validation performance than both the standard likelihood-based loss and the sigmoid-smoothed C-index, and is therefore selected as the optimization objective.

\section{Experiments}

\subsection{Evaluation metrics}
Evaluating survival models requires metrics that account for censoring and
time-to-event outcomes. The most widely used measure is \emph{Harrell's
concordance index (C-index)}, which quantifies a model ability to correctly rank event times. Let
$\mathcal{J}$ denote the set of all pairs $(i,j)$ with $\delta_{i}=1$ and
$t_{i} < t_{j}$. The C-index is computed as
\begin{equation}
C=\frac{\sum_{(i,j)\in \mathcal{J}}\mathbf{1}\! \left[  \hat{t}_{i}<\hat{t}_{j}\right]  }{|\mathcal{J}|},
\end{equation}
where $\hat{t}_{i}$ and $\hat{t}_{j}$ are the expected event times approximations derived from the predicted survival function. A
C-index of $1$ indicates perfect ranking, while a value of $0.5$ corresponds
to random ordering.

Another important metric is the \emph{Integrated Brier Score (IBS)}, a
time-dependent measure that combines discrimination and calibration. For a
given time interval $[0,t_{\max}]$, it is defined as
\begin{equation}
IBS=\frac{1}{t_{\max}}\int_{0}^{t_{\max}}\! \mathbb{E}\! \left[  \left(
\Delta_{\text{new}}(t)-\hat{S}(t\mid \mathbf{x}_{\text{new}})\right)
^{2}\right]  dt.
\end{equation}
where $\Delta_{\text{new}}(t)=\mathbf{1}(T_{\text{new}}>t)$ is the true
survival status at $t$ and $\hat{S}(t\mid \mathbf{x}_{\text{new}})$ is the
predicted survival probability. Lower values of the IBS indicate better
predictive accuracy.

A complementary time-dependent discrimination metric is the
\emph{cumulative/dynamic area under the ROC curve (AUC)}.
At a given time $t$, the time-dependent AUC evaluates the ability of a
model to distinguish individuals who experience the event by time $t$ from
those who remain event-free at $t$.
Formally, it is defined as the probability that, for a randomly selected pair
consisting of one individual with $T \le t$ and one individual with $T > t$,
the model assigns a higher risk score to the former.

In practice, we compute the cumulative/dynamic AUC using the predicted survival
probabilities $\widehat{S}(t \mid \mathbf{x})$, with appropriate handling of censoring.
To obtain a single summary statistic, we report the time-integrated AUC by
averaging the time-dependent AUC over the evaluation horizon.
Higher values of the AUC indicate better discriminative
performance.

\subsection{Datasets}
We evaluate the proposed methods on a collection of 10 real-world survival
datasets. The list of datasets and their main characteristics are summarized
in Table~\ref{tab:datasets} (Appendix~\ref{sec:datasets}).

All datasets are preprocessed in a unified manner prior to training.
Numerical features are standardized, while categorical features are encoded
using a one-hot scheme.
Missing values are imputed with the mean for numerical features and with an
additional category for categorical features.
Features that are missing for all samples are removed from the dataset.

\subsection{Methodology}
All real-world datasets are split into training, validation, and test sets,
with stratification based on censoring indicators.
The experimental procedure consists of two stages.
The design of the experimental protocol follows the methodology commonly used
in modern tabular neural network studies, such as TabR
\cite{gorishniy2024tabr} and TabM \cite{Gorishniy2024TabMAT}. This protocol
is particularly appropriate for neural models whose training is
computationally expensive and whose performance is sensitive to
hyperparameter choices. The same protocol is also applied to all competing
methods, for which hyperparameter tuning is similarly important. For each
dataset, we use a fixed train--validation--test split and repeat training and
evaluation with different random seeds. Hyperparameters are selected once for
each method within the fixed train--validation part of each dataset, using the
validation set, and are then kept fixed across repeated runs rather than
re-optimized for each random seed. Thus, the protocol does not use nested
cross-validation. Nevertheless, all methods are evaluated under identical
conditions, and the held-out test set is never used for model selection or
hyperparameter tuning, preventing adaptation to the test portion of a
particular split. This setup provides reliable comparisons while remaining
feasible for expensive neural baselines. Moreover, most datasets contain at least 500
observations after preprocessing, and the five largest datasets have held-out
test sets ranging from 572 to 2732 samples (Table~\ref{tab:datasets}), making
the use of held-out test evaluation appropriate for the benchmark.

In the first stage, hyperparameters are tuned using the Tree-structured
Parzen Estimator (TPE) sampler implemented in Optuna \cite{akiba2019optuna}.
The optimal hyperparameter configuration is selected by maximizing the
C-index on the validation set.
The number of optimization iterations depends on the number of hyperparameters
of a given method and ranges from 50 to 200.
At this stage, the validation set effectively serves as a test set, while
models are trained on the training set.
For methods employing early stopping, the validation set is sampled as
25\% of the training data.

The second stage corresponds to the final evaluation.
Depending on the method, the validation set is either used for early stopping
or merged with the training set (as in the case of Random Survival Forests).
All evaluation metrics are computed on the held-out test set.
At each experimental run, models are ranked according to their performance,
separately for the C-index, IBS, and time-dependent AUC.
The reported results correspond to the average metric values and average ranks
computed over 20 independent runs.

\subsection{Results on real-world datasets}

\begin{figure}[t]
    \centering
    \includegraphics[width=0.7\columnwidth]{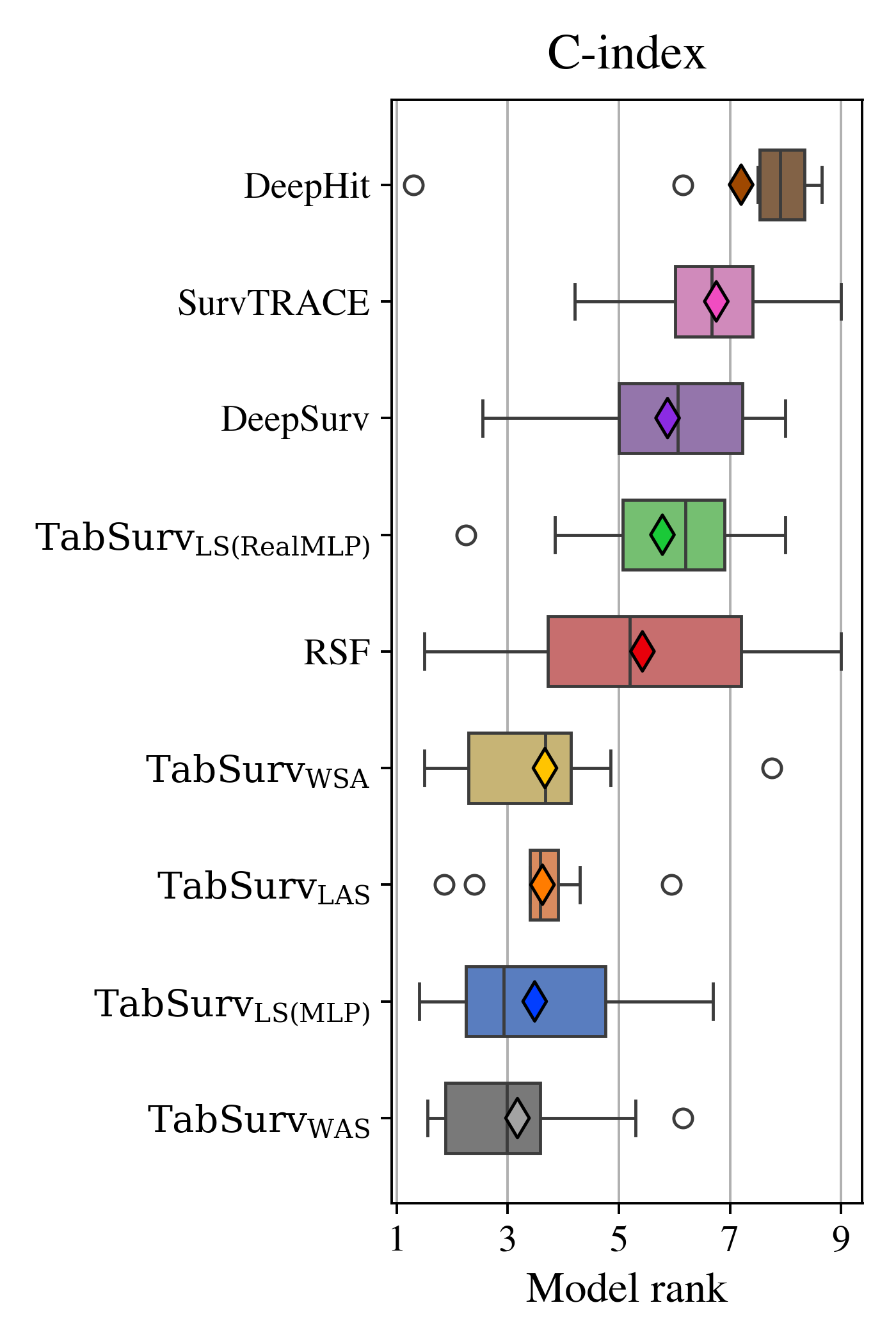}
    \caption{Comparison of model performance across all datasets based on the
    C-index. Diamonds indicate mean ranks.}
    \label{fig:ranks_cindex}
\end{figure}

\begin{figure}[t]
    \centering
    \includegraphics[width=0.7\columnwidth]{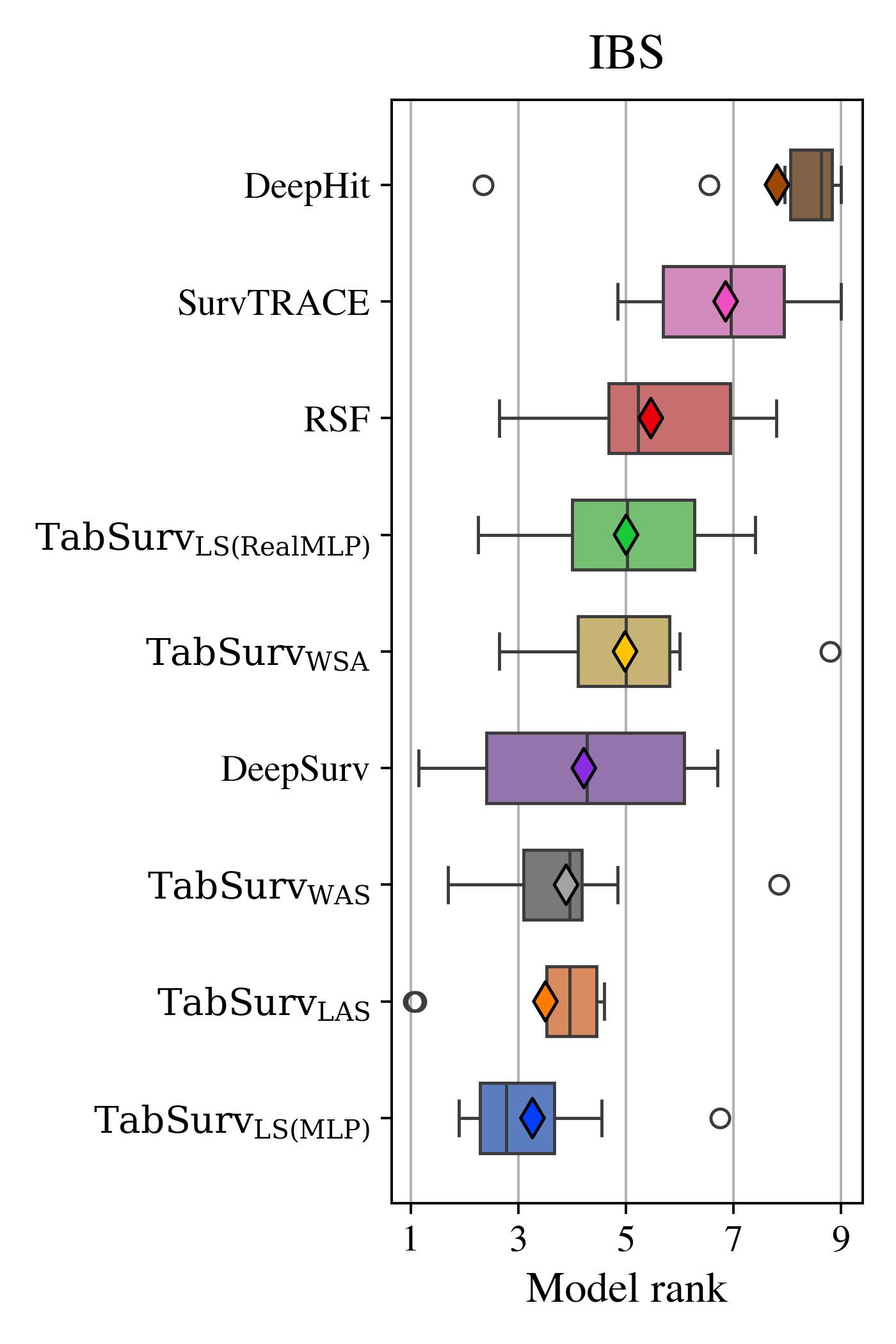}
    \caption{Comparison of model performance across all datasets based on the
    Integrated Brier Score (IBS). Diamonds indicate mean ranks.}
    \label{fig:ranks_ibs}
\end{figure}

\begin{figure}[t]
    \centering
    \includegraphics[width=0.7\columnwidth]{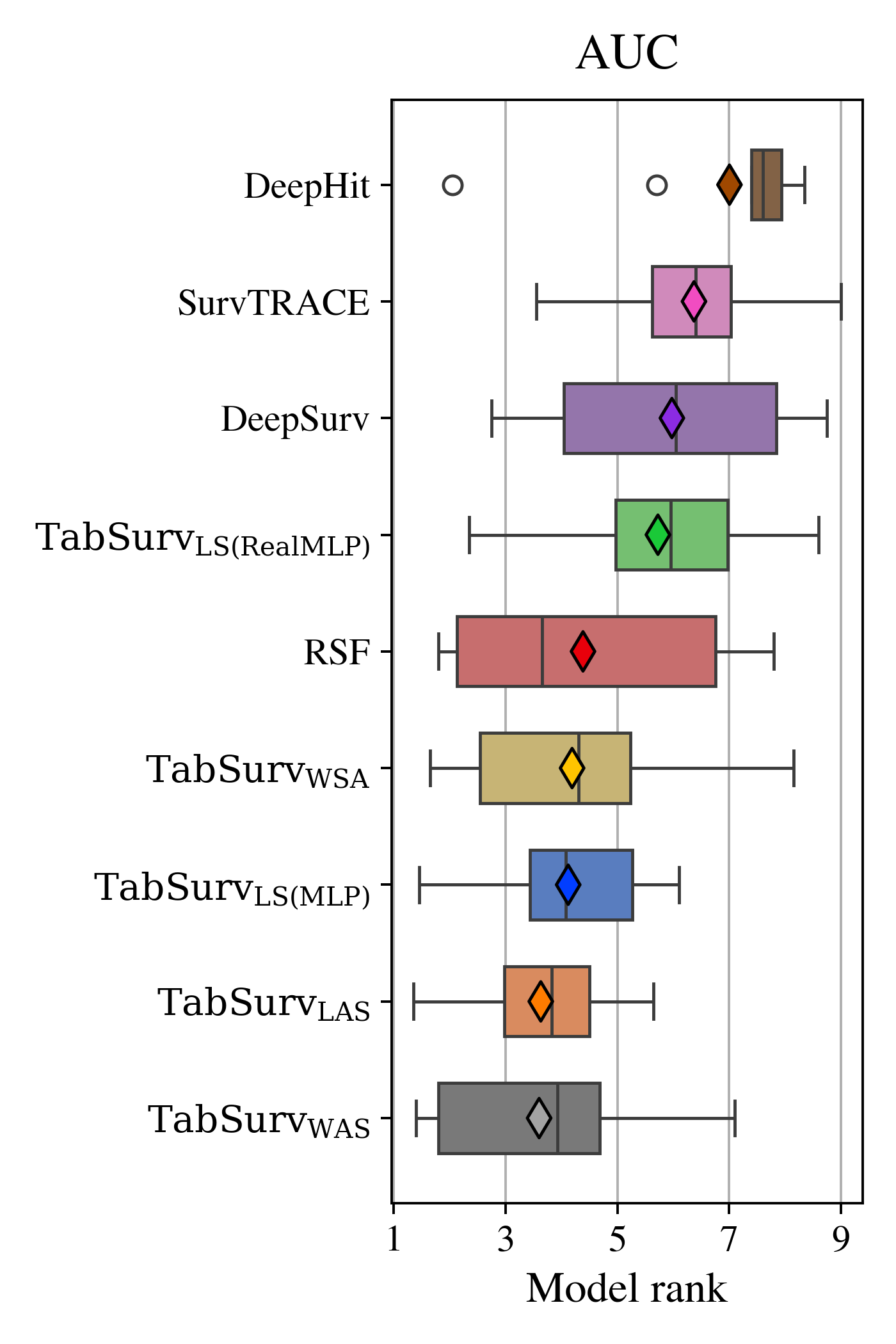}
    \caption{Comparison of model performance across all datasets based on the
    time-dependent AUC. Diamonds indicate mean ranks.}
    \label{fig:ranks_auc}
\end{figure}

The results are summarized using rank-based comparisons across all datasets.
For each method, an average rank is computed on each dataset, and the
distribution of these average ranks is visualized using boxplots.
Figures~\ref{fig:ranks_cindex}, \ref{fig:ranks_ibs}, and \ref{fig:ranks_auc}
report these rank distributions for the C-index, IBS, and time-dependent AUC,
respectively. Detailed result tables for each dataset are provided in Appendix~\ref{res_tables}.
Methods are ordered according to their mean rank aggregated over all datasets,
with lower ranks indicating better overall performance.
In all figures, the evaluated methods are denoted as follows. 
The proposed models are referred to as TabSurv, with a subscript indicating the specific architectural variant (LAS, WSA, or WAS). 
For the LS architecture, the employed backbone is additionally specified in parentheses (MLP or RealMLP). 
The Random Survival Forest baseline is abbreviated as \textit{RSF}. 
All remaining methods are denoted by their full names without abbreviations.

Across all datasets and evaluation metrics, the best overall performance is achieved by the proposed TabSurv\textsubscript{WAS}, TabSurv\textsubscript{LAS}, and TabSurv\textsubscript{LS(MLP)} models, while all existing baseline methods consistently yield inferior results with respect to C-index, IBS, and time-dependent AUC.
The results indicate that employing RealMLP as a backbone without additional architectural adaptations leads to worse performance compared to standard MLPs. Furthermore, the WSA variant underperforms relative to WAS, which is consistent with theoretical considerations regarding noise smoothing properties inherent to the WAS formulation.
The LS(MLP) and LAS architectures achieve the lowest Integrated Brier Score, suggesting that these methods provide the most accurate estimates of the survival function shape. In contrast, the WAS variant attains superior C-index and time-dependent AUC values, indicating more accurate ranking of expected event times. This behavior may be attributed to the comparatively simpler optimization of neural networks predicting Weibull distribution parameters, as opposed to models that predict a large number of logits.

\subsection{Additional analysis}

The empirical results also provide practical guidance on which TabSurv variant
to use. The LS(MLP), LAS, and WAS variants are the most accurate overall, but
they are preferable in different regimes. The WAS model is attractive when a
compact parametric description of event times is adequate, because the
two-parameter Weibull head acts as a strong regularizer. This reduction in
variance can improve ranking metrics such as the C-index even when the exact
Weibull assumption is imperfect. However, the simulation study in
Appendix~\ref{sec:simulation} shows that non-parametric variants are better
suited to genuinely multimodal event-time distributions. Between LS(MLP) and
LAS, the former is a more economical choice for smaller datasets, whereas LAS
is preferable on larger datasets, where its parameter-efficient ensemble can
benefit from more data.

The framework is not tied to a specific backbone or distributional family.
In this paper we instantiate it with MLP, TabM-like, and RealMLP-style
backbones, observing that the MLP and TabM-like variants are the strongest in
the considered survival setting. RealMLP performs worse without the kind of
large-scale pretraining used in its original tabular learning context, which
suggests that survival-specific adaptation remains important. Extending
TabSurv to retrieval-based models such as TabR, or to foundation-style
tabular models such as TabPFN and MotherNet, is conceptually possible but
requires additional work on survival-compatible output heads and training
objectives. Similarly, the Weibull head is only one parametric instance; other
families can be substituted by changing the distribution used to map predicted
parameters to survival probabilities.

To assess whether the gains are merely a consequence of using ensembles, we
compare TabSurv with independently ensembled DeepSurv and DeepHit baselines in
Appendix~\ref{sec:ensemble_ablation}. The results indicate that external
ensembling alone does not explain the performance of TabSurv: LS(MLP) already
outperforms DeepSurv and remains competitive with its ensemble, while the
independent LS(MLP) ensemble is practically equivalent to LAS but requires
substantially more computation. We also compare TabSurv with GBDT-based
survival models based on XGBoost survival variants \cite{Barnwal-etal-22}.
Table~\ref{tab:xgbse_comparison} reports Bayesian probabilities that
TabSurv\textsubscript{LS(MLP)} is better, practically equivalent, or worse
than two XGBSE variants. TabSurv is either competitive or favorable across the
considered metrics, with particularly strong IBS performance against the
stacked Weibull variant.

\begin{table*}[t]
\centering
\small
\begin{tabular}{llccc}
\toprule
Baseline & Metric & Better & ROPE & Worse \\
\midrule
XGBSEKaplanNeighbors & C-index & 0.567 & 0.254 & 0.179 \\
XGBSEKaplanNeighbors & IBS & 0.293 & 0.605 & 0.102 \\
XGBSEKaplanNeighbors & AUC & 0.796 & 0.140 & 0.064 \\
XGBSEStackedWeibull & C-index & 0.254 & 0.742 & 0.004 \\
XGBSEStackedWeibull & IBS & 0.968 & 0.032 & 0.000 \\
XGBSEStackedWeibull & AUC & 0.427 & 0.417 & 0.156 \\
\bottomrule
\end{tabular}
\caption{Bayesian comparison between TabSurv\textsubscript{LS(MLP)} and
XGBSE-based survival baselines. For C-index and AUC, ``Better'' denotes
$P(\mathrm{TabSurv}>\mathrm{XGBSE})$; for IBS, it denotes
$P(\mathrm{TabSurv}<\mathrm{XGBSE})$. The ROPE width is 0.005.}
\label{tab:xgbse_comparison}
\end{table*}

\section{Conclusion}

We proposed a unified neural framework for tabular survival analysis based on the Survival Histogram Loss and introduced a family of models, TabSurv, that differ in their survival distribution parameterization and aggregation strategies. The framework is architecture-agnostic and supports heterogeneous tabular features through flexible embedding mechanisms.

We evaluated the proposed methods on a diverse set of real-world survival datasets using a rigorous experimental protocol with hyperparameter optimization and repeated runs. Performance was assessed using complementary metrics, C-index, Integrated Brier Score (IBS), and time-dependent AUC, capturing both ranking quality and survival function accuracy.

Across all datasets, TabSurv\textsubscript{WAS}, TabSurv\textsubscript{LAS}, and TabSurv\textsubscript{LS(MLP)} consistently outperformed existing baselines. LAS and LS(MLP) achieved the lowest IBS values, indicating more accurate estimation of survival function shapes, while WAS showed superior performance in terms of C-index and time-dependent AUC, suggesting more reliable event-time ordering. The results also indicate that more complex backbones, such as RealMLP, do not necessarily improve performance without additional adaptation, and that WAS is preferable to WSA, in line with theoretical considerations on noise smoothing.

Overall, the proposed framework establishes a strong baseline for neural survival modeling on tabular data and offers a flexible foundation for future extensions, including alternative distributional assumptions and more advanced backbone architectures.

\section*{Impact Statement}

This paper presents work whose goal is to advance the field of Machine
Learning. The proposed methods are intended for general-purpose survival modeling and decision support. As with all predictive models applied to sensitive domains, careful validation and responsible use are required.

%In the unusual situation where you want a paper to appear in the
%references without citing it in the main text, use \nocite
%\nocite{langley00}

\bibliographystyle{plain}
\bibliography{Attention,Survival_analysis,TabTrans,Transformer,Classif_bib}

\newpage \appendix
\onecolumn

\section{Datasets}
\label{sec:datasets}

All statistics reported in Table~\ref{tab:datasets} correspond to the
datasets after preprocessing. Additional remarks on datasets:
\begin{itemize}
    \item \textbf{SEER.} The processed version of the SEER dataset is obtained from \url{https://dx.doi.org/10.21227/a9qy-ph35}, where records from multiple years are consistently merged.
    The official documentation for the raw SEER data is available at
    \url{https://seer.cancer.gov/data-software/documentation/seerstat/nov2017/}.
    \item \textbf{TCGA-GBM.} This dataset is derived from the TCGA Pan-Cancer Atlas \cite{liu2018integrated} and includes only patients diagnosed with Glioblastoma Multiforme (GBM).
    \item For the METABRIC dataset, we use both clinical features and
    \emph{mRNA expression z-scores}, whereas for all other datasets only
    clinical features are employed.
\end{itemize}

\begin{table}[H]
    \centering
    \begin{tabular}{|c|c|c|c|c|c|c|}
        \hline
        Name & \multicolumn{3}{c|}{Number of samples} & \multicolumn{2}{c|}{Number of features} & \multirow{2}{*}{Event rate} \\
        \cline{2-6}
        & Train & Validation & Test & Numeric & Categorical & \\
        \hline
        METABRIC \cite{curtis2012genomic} & 999 & 333 & 572 & 494 & 75 & 42\% \\
        \hline
        SEER & 2112 & 704 & 1208 & 4 & 21 & 15\% \\
        \hline
        SUPPORT \cite{connors1995controlled} & 4779 & 1594 & 2732 & 37 & 30 & 68\% \\
        \hline
        TCGA-GBM \cite{liu2018integrated} & 312 & 105 & 179 & 5 & 20 & 82\% \\
        \hline
        Rotterdam \cite{royston2013external} & 1565 & 522 & 895 & 5 & 6 & 43\% \\
        \hline
        FLC \cite{dispenzieri2012use} & 4133 & 1378 & 2363 & 4 & 34 & 28\% \\
        \hline
        GBSG2 \cite{schumacher1994randomized} & 360 & 120 & 206 & 5 & 4 & 44\% \\
        \hline
        WHAS500 \cite{hosmer2008applied} & 262 & 88 & 150 & 6 & 8 & 43\% \\
        \hline
        Lung \cite{loprinzi1994prospective} & 119 & 40 & 69 & 3 & 4 & 72\% \\
        \hline
        PBC \cite{fleming2013counting} & 219 & 73 & 126 & 11 & 15 & 39\% \\
        \hline
    \end{tabular}
    \caption{Summary of the datasets used in the experiments.}
    \label{tab:datasets}
\end{table}

\section{Experiment setup}

During the experiments, SurvTRACE and several TabSurv models (LAS, WAS, WSA) were executed on an NVIDIA Tesla V100 GPU with 16 GB of memory using CUDA 12.1. Other models (TabSurv\textsubscript{LS}, DeepHit, DeepSurv, and RSF) were trained on a CPU using up to eight threads simultaneously.

\subsection{Tuned hyperparameters}

The hyperparameters were tuned using the Optuna
framework \cite{akiba2019optuna} for each evaluated model, with the following hyperparameter domains:

\begin{itemize}
    \item \textbf{TabSurv\textsubscript{LS(MLP)}} %(200 optimization trials)
    \begin{itemize}
        \item Number of layers: 1 to 4
        \item Number of hidden units per layer: 128, 256, or 512
        \item Activation function: ReLU, SiLU, or SELU
        \item Loss parameter $r$: 1 to 5
        \item Learning rate: $\alpha \sim \mathrm{LogUniform}[10^{-4}, 5 \cdot 10^{-3}]$
        \item Batch size: 32, 64, or 96
        \item Use of layer normalization between layers
        \item Number of bins in Piecewise Linear Embeddings: 32, 48, or 64
        \item Use of activation function in numerical embeddings
        \item Dimensionality of numerical embeddings: 8, 12, or 16
    \end{itemize}

    \item \textbf{TabSurv\textsubscript{LAS}} %(200 optimization trials)
    \begin{itemize}
        \item Number of blocks: 1 to 3
        \item Number of hidden units per block: 128, 256, or 512
        \item Number of neural networks in the ensemble: 8, 16, or 32
        \item Activation function: ReLU, SiLU, or SELU
        \item Loss parameter $r$: 1 to 5
        \item Learning rate: $\alpha \sim \mathrm{LogUniform}[10^{-4}, 5 \cdot 10^{-3}]$
        \item Batch size: 32, 64, or 96
        \item Dropout rate: $\theta \sim \mathrm{LogUniform}[10^{-2}, 10^{-1}]$
        \item Number of bins in Piecewise Linear Embeddings: 32, 48, or 64
        \item Use of activation function in numerical embeddings
        \item Dimensionality of numerical embeddings: 8, 12, or 16
    \end{itemize}

    \item \textbf{TabSurv\textsubscript{LS(RealMLP)}} %(200 optimization trials)
    \begin{itemize}
        \item Number of layers: 1 to 4
        \item Number of hidden units per layer: 128, 256, or 512
        \item Activation function: Mish or SELU
        \item Loss parameter $r$: 1 to 5
        \item Learning rate: $\alpha \sim \mathrm{LogUniform}[10^{-4}, 5 \cdot 10^{-3}]$
        \item Batch size: 32, 64, or 96
        \item Number of frequencies in Periodic Bias Linear DenseNet embeddings: 32, 48, or 64
        \item Frequency scale in numerical embeddings: $\rho \sim \mathrm{LogUniform}[10^{-2}, 1]$
        \item Dimensionality of numerical embeddings: 8, 12, or 16
    \end{itemize}

    \item \textbf{TabSurv\textsubscript{WAS}} %(120 optimization trials)
    \begin{itemize}
        \item Number of blocks: 1 to 3
        \item Number of hidden units per block: 64, 128, or 256
        \item Number of neural networks in the ensemble: 8, 16, or 32
        \item Activation function: ReLU or SELU
        \item Loss parameter $r$: 1, 3, or 5 bins
        \item Learning rate: $\alpha \sim \mathrm{LogUniform}[10^{-4}, 5 \cdot 10^{-3}]$
        \item Batch size: 32, 64, or 96
        \item Number of bins in Piecewise Linear Embeddings: 32, 48, or 64
        \item Use of activation function in numerical embeddings
        \item Dimensionality of numerical embeddings: 8, 12, or 16
    \end{itemize}

    \item \textbf{TabSurv\textsubscript{WSA}} %(120 optimization trials)  
    Hyperparameters are identical to those of \textbf{TabSurv\textsubscript{WSA}}.

    \item \textbf{SurvTRACE} (100 optimization trials)
    \begin{itemize}
        \item Number of hidden layers: 1 to 4
        \item Transformer hidden size: 128, 256, or 512
        \item Transformer intermediate size: 128, 256, or 512
        \item Number of attention heads: 2, 4, or 8
        \item Learning rate: $\alpha \sim \mathrm{LogUniform}[10^{-4}, 5 \cdot 10^{-3}]$
        \item Dropout rate: $\theta \sim \mathrm{LogUniform}[10^{-2}, 10^{-1}]$
        \item Batch size: 32 or 64
        \item Weight decay: $\gamma \sim \mathrm{LogUniform}[10^{-4}, 10^{-1}]$
    \end{itemize}

    \item \textbf{DeepHit} %(150 optimization trials)
    \begin{itemize}
        \item Number of hidden layers: 1 to 4
        \item Number of hidden units per layer: 128, 256, or 512
        \item Activation function: ReLU or Tanh
        \item Batch size: 64, 128, or 256
        \item Learning rate: $\alpha_{\mathrm{step}} \sim \mathrm{LogUniform}[10^{-4}, 5 \cdot 10^{-3}]$
        \item Dropout rate: $\theta \sim \mathrm{LogUniform}[10^{-2}, 10^{-1}]$
        \item Use of batch normalization between layers
        \item Loss weighting coefficient $\alpha$ in DeepHit: $\alpha \sim \mathrm{U}[0, 1]$
        \item Smoothing parameter $\sigma$ in DeepHit loss: $10^{-3},\ 10^{-2},\ 10^{-1}$, or $1$
    \end{itemize}

    \item \textbf{DeepSurv} %(100 optimization trials)
    \begin{itemize}
        \item Number of hidden layers: 1 to 4
        \item Number of hidden units per layer: 128, 256, or 512
        \item Activation function: ReLU or Tanh
        \item Batch size: 64, 128, or 256
        \item Learning rate: $\alpha \sim \mathrm{LogUniform}[10^{-4}, 5 \cdot 10^{-3}]$
        \item Dropout rate: $\theta \sim \mathrm{LogUniform}[10^{-2}, 10^{-1}]$
        \item Use of batch normalization between layers
    \end{itemize}

    \item \textbf{Random Survival Forest} %(50 optimization trials)
    \begin{itemize}
        \item Number of trees in the ensemble: 20, 50, 100, or 200
        \item Maximum tree depth: 3, 5, or unlimited
        \item Minimum number of samples required to split a node: 2 samples, $1\%$ of the dataset size, or $5\%$
        \item Number of features considered at each split: $\sqrt{d}$ or $\log_2 d$, where $d$ denotes the number of features
    \end{itemize}
\end{itemize}

\section{Additional ablations and statistical comparisons}
\label{sec:additional_ablations}

We complement the rank-based summary in the main text with Bayesian
comparisons following \cite{benavoli2017time}. For each dataset and method,
the metric value is averaged over 20 runs with different random seeds. The
Bayesian comparison is then performed across datasets with a region of
practical equivalence (ROPE) of 0.005. For the C-index and AUC, larger values
are better; for IBS, smaller values are better.

%\begin{figure}[H]
%    \centering
%    \includegraphics[width=\linewidth,height=0.78\textheight,keepaspectratio]{bayesian_c-index.pdf}
%    \caption{Bayesian pairwise comparison of methods by C-index.}
%    \label{fig:bayesian_cindex}
%\end{figure}
%
%\begin{figure}[H]
%    \centering
%    \includegraphics[width=\linewidth,height=0.78\textheight,keepaspectratio]{bayesian_ibs.pdf}
%    \caption{Bayesian pairwise comparison of methods by IBS.}
%    \label{fig:bayesian_ibs}
%\end{figure}

\begin{figure}[p]
    \centering
    \begin{minipage}[b]{0.48\textwidth}
        \centering
        \includegraphics[width=\linewidth,keepaspectratio]{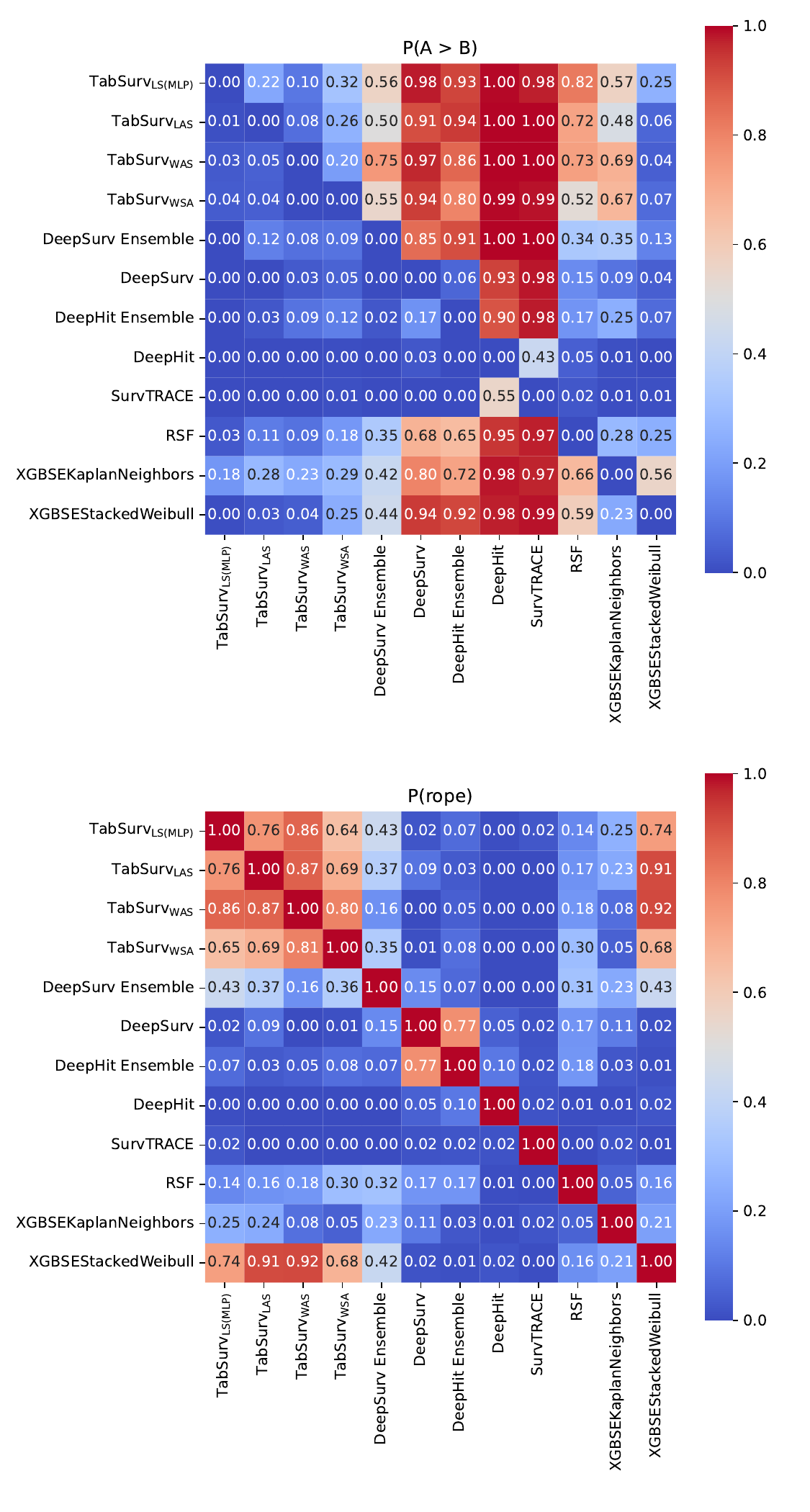}
        \subcaption{Bayesian pairwise comparison of methods by C‑index.}
        \label{fig:bayesian_cindex}
    \end{minipage}
    \hfill
    \begin{minipage}[b]{0.48\textwidth}
        \centering
        \includegraphics[width=\linewidth,keepaspectratio]{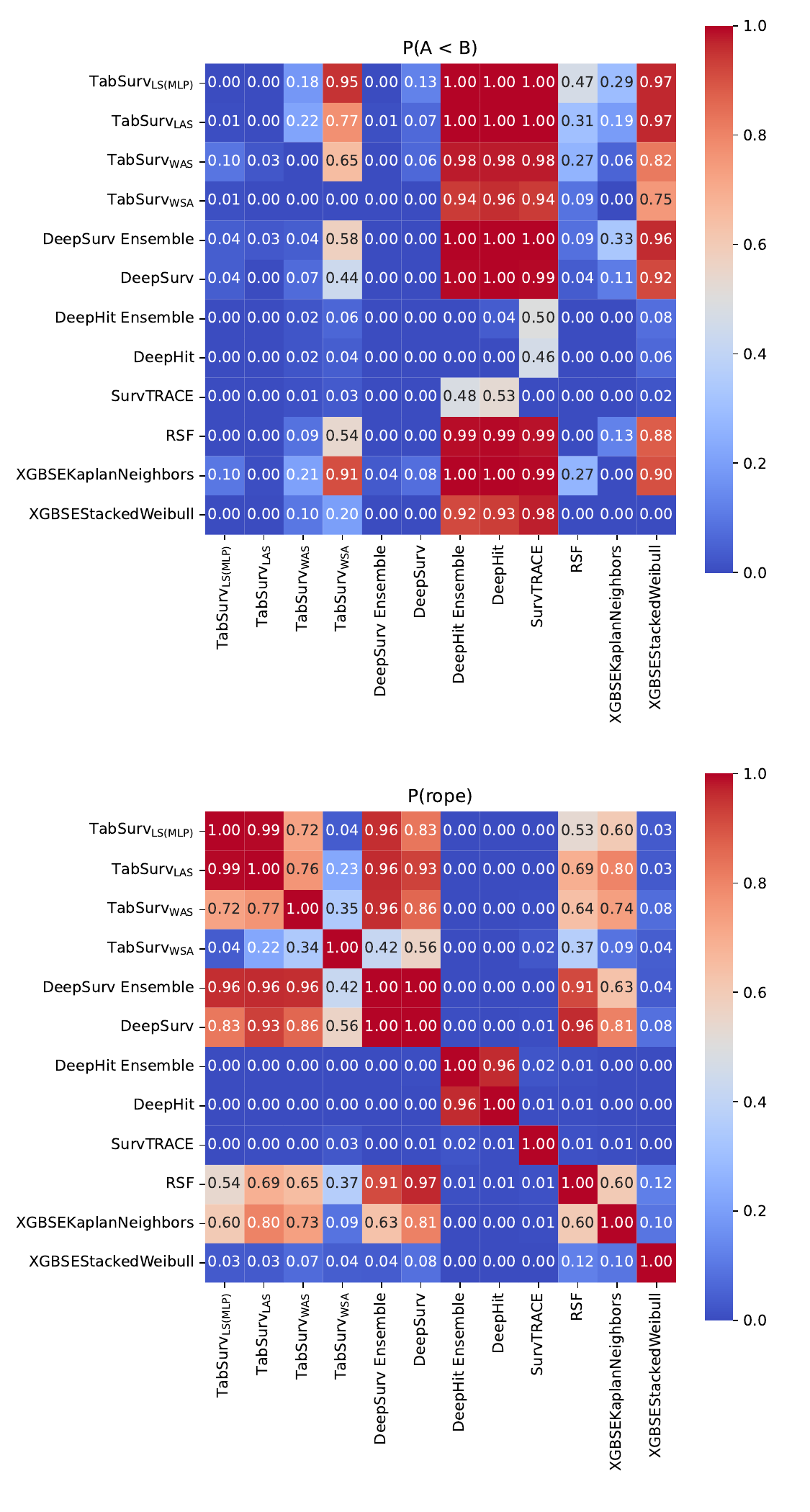}
        \subcaption{Bayesian pairwise comparison of methods by IBS.}
        \label{fig:bayesian_ibs}
    \end{minipage}
    \caption{Bayesian pairwise comparisons of methods: (a) by C‑index; (b) by IBS.}
    \label{fig:bayesian_combined}
\end{figure}

\subsection{Component ablation}
\label{sec:component_ablation}

To isolate the contribution of the main components, we start from a baseline
MLP and progressively add numerical embeddings (Emb), ensembling (Ens), the
SurvHL loss, and the Weibull head. Table~\ref{tab:ablation_cindex} reports
the posterior probability that the model in the row outperforms the model in
the column in terms of the C-index. The results show that SurvHL, embeddings,
and ensembling are most effective when combined: the full non-parametric
variant and the Weibull variant both outperform the baseline and most partial
configurations with high posterior probability. The Weibull head adds a
compact parametric bias that further improves C-index over several
non-parametric variants.

\begin{table}
\centering
\small
\begin{tabular}{lrrrrrrr}
\toprule
$P(A>B)$, C-index & MLP & Emb & Ens & Emb+Ens & SurvHL & SurvHL+Ens & Emb+SurvHL \\
\midrule
Emb+SurvHL+Ens+Weib & 0.914 & 0.947 & 0.734 & 0.505 & 0.900 & 0.743 & 0.862 \\
Emb+SurvHL+Ens & 0.959 & 0.968 & 0.709 & 0.748 & 0.883 & 0.265 & 0.755 \\
SurvHL+Ens & 0.785 & 0.993 & 0.596 & 0.742 & 0.388 & -- & 0.654 \\
SurvHL & 0.249 & 0.892 & 0.140 & 0.281 & -- & 0.123 & 0.180 \\
\bottomrule
\end{tabular}
\caption{Bayesian ablation of TabSurv components. Rows correspond to model
$A$, columns to model $B$.}
\label{tab:ablation_cindex}
\end{table}

%\begin{figure}[H]
%    \centering
%    \includegraphics[width=\linewidth,height=0.78\textheight,keepaspectratio]{ablation_bayesian_c-index.pdf}
%    \caption{Full Bayesian ablation comparison by C-index.}
%    \label{fig:ablation_bayesian_cindex}
%\end{figure}
%
%\begin{figure}[H]
%    \centering
%    \includegraphics[width=\linewidth,height=0.78\textheight,keepaspectratio]{ablation_bayesian_ibs.pdf}
%    \caption{Full Bayesian ablation comparison by IBS.}
%    \label{fig:ablation_bayesian_ibs}
%\end{figure}

\begin{figure}
    \centering
    \begin{minipage}[b]{0.48\textwidth}
        \centering
        \includegraphics[width=\linewidth,keepaspectratio]{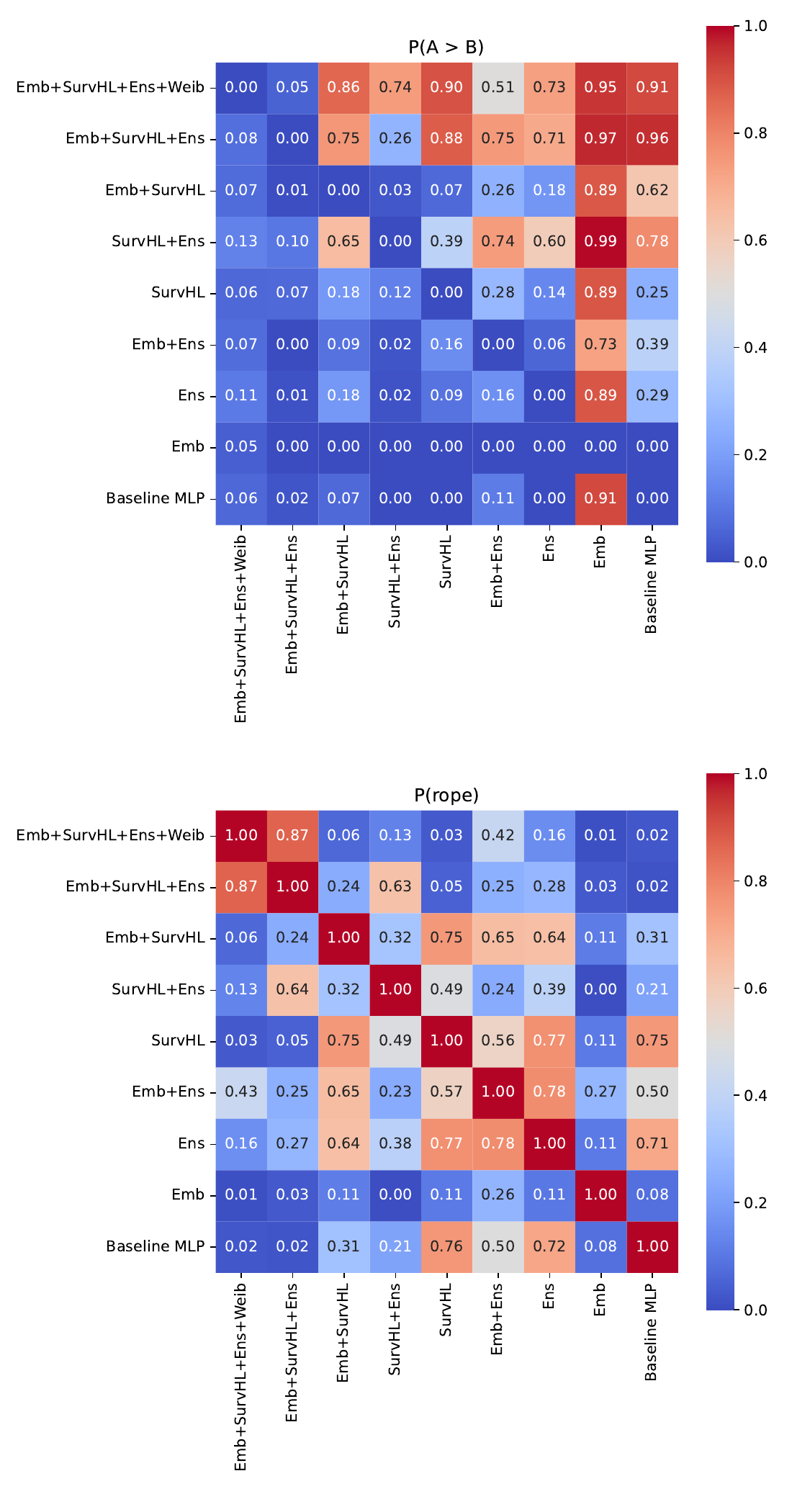}
        \subcaption{Full Bayesian ablation comparison by C‑index.}
        \label{fig:ablation_bayesian_cindex}
    \end{minipage}
    \hfill
    \begin{minipage}[b]{0.48\textwidth}
        \centering
        \includegraphics[width=\linewidth,keepaspectratio]{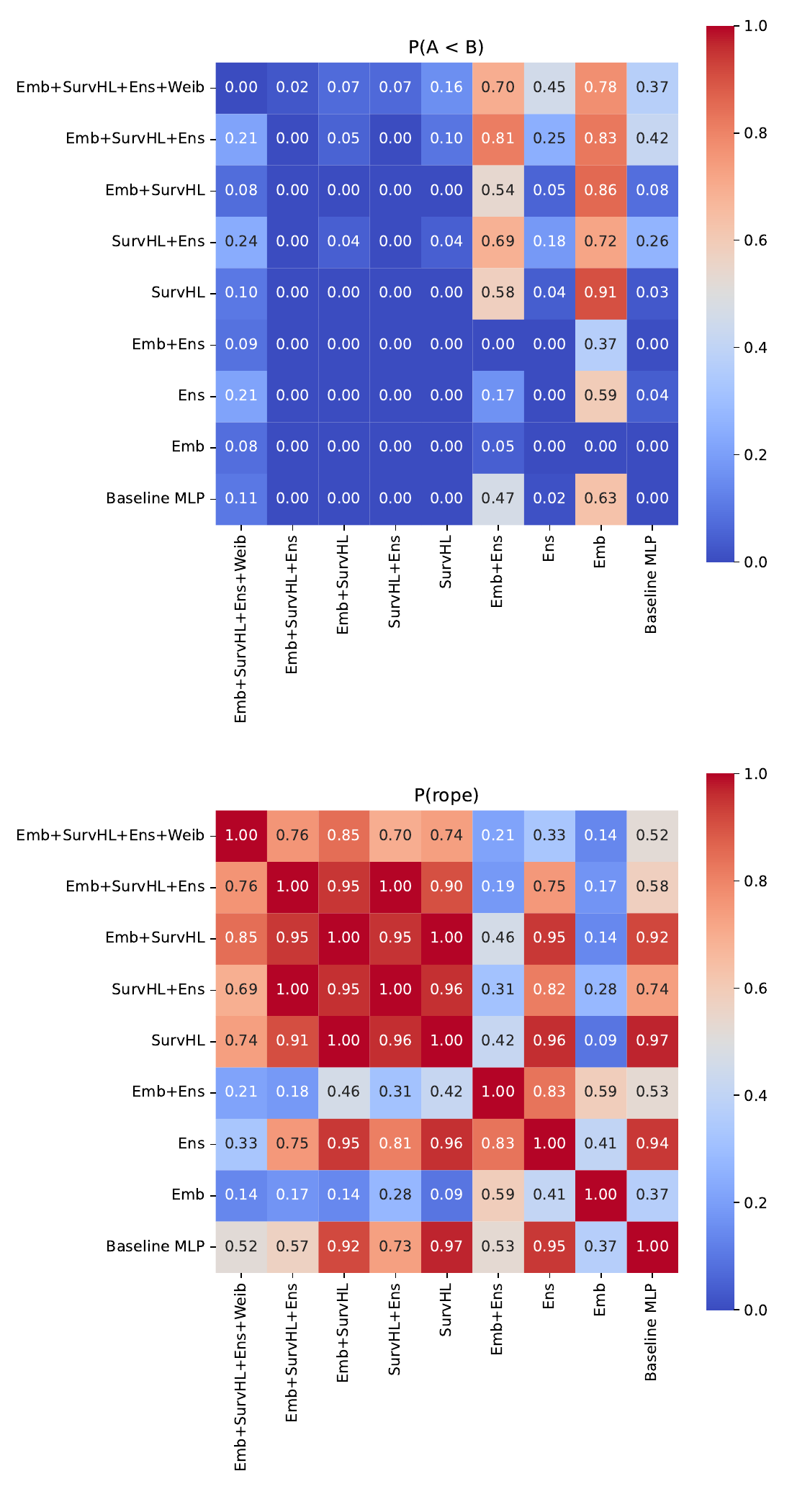}
        \subcaption{Full Bayesian ablation comparison by IBS.}
        \label{fig:ablation_bayesian_ibs}
    \end{minipage}
    \caption{Full Bayesian ablation comparisons: (a) by C‑index; (b) by IBS.}
    \label{fig:ablation_bayesian_combined}
\end{figure}

\begin{figure}
    \centering
    \includegraphics[width=0.6\linewidth,keepaspectratio]{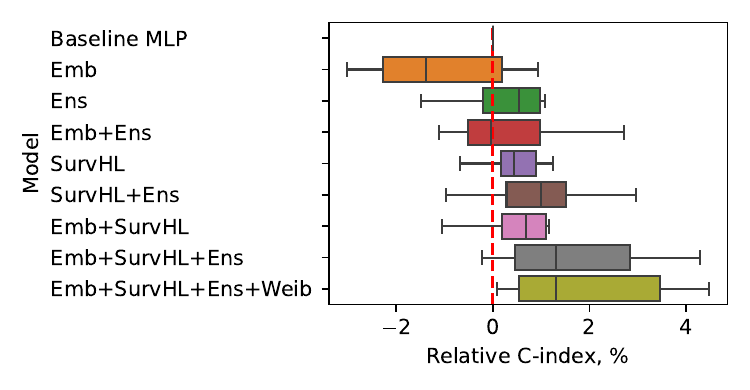}
    \caption{Relative component ablation results by C-index.}
    \label{fig:ablation_relative_cindex}
\end{figure}

\begin{figure}
    \centering
    \includegraphics[width=0.6\linewidth,keepaspectratio]{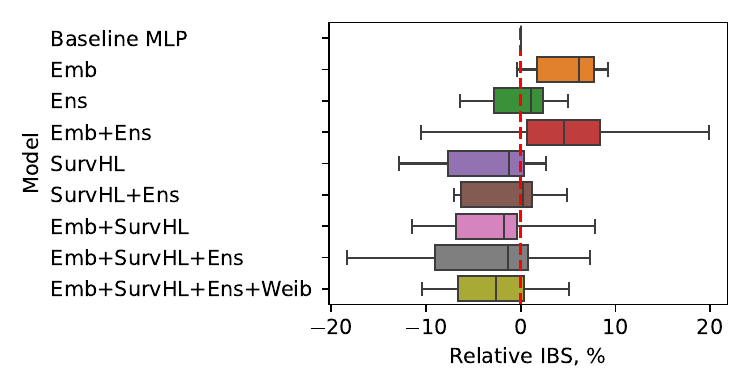}
    \caption{Relative component ablation results by IBS.}
    \label{fig:ablation_relative_ibs}
\end{figure}

\subsection{Sensitivity of SurvHL}
\label{sec:survhl_sensitivity}

\begin{figure}
    \centering
    \includegraphics[width=0.5\linewidth]{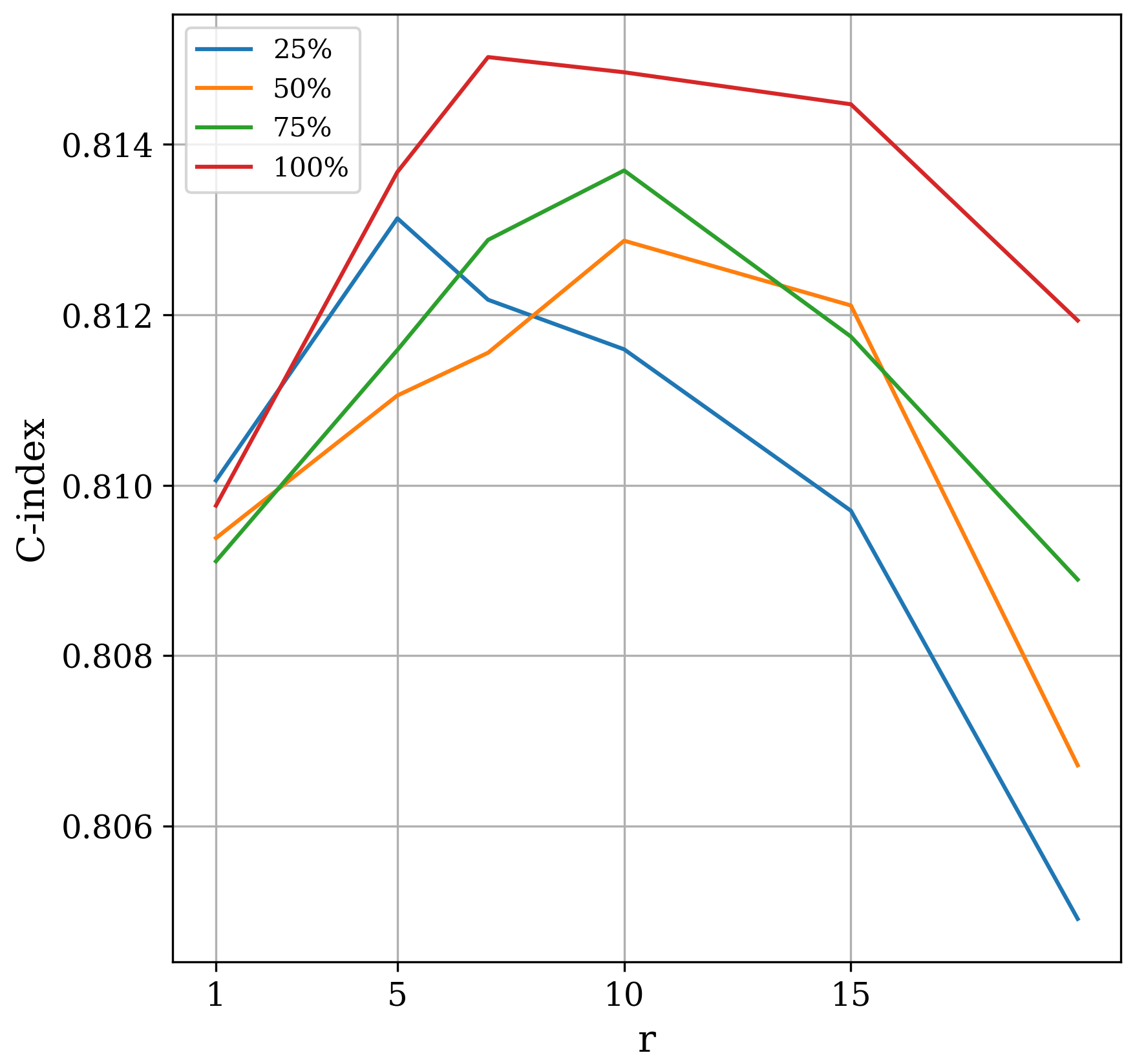}
    \caption{Sensitivity of the METABRIC C-index to the SurvHL smoothing
    parameter and discretization grid.}
    \label{fig:loss_parameters_metabric}
\end{figure}

SurvHL introduces a smoothing parameter $r$ and depends on the discretization
grid used to represent the event-time distribution. Table~\ref{tab:survhl_grid}
and Figure~\ref{fig:loss_parameters_metabric} report C-index values on the
METABRIC dataset for several grid sizes and values of $r$. For each grid,
there is a preferred smoothing level, and the best value is obtained with the
full grid. At the same time, the total variation across this range of
settings is small, indicating that SurvHL is not overly sensitive to these
hyperparameters.

\begin{table}
\centering
\small
\begin{tabular}{lrrrrrr}
\toprule
Grid & $r=1$ & $r=5$ & $r=7$ & $r=10$ & $r=15$ & $r=20$ \\
\midrule
25\% & 0.81006 & 0.81313 & 0.81218 & 0.81160 & 0.80970 & 0.80491 \\
50\% & 0.80938 & 0.81105 & 0.81156 & 0.81287 & 0.81211 & 0.80671 \\
75\% & 0.80911 & 0.81159 & 0.81288 & 0.81369 & 0.81175 & 0.80889 \\
100\% & 0.80976 & 0.81368 & 0.81503 & 0.81485 & 0.81447 & 0.81193 \\
\bottomrule
\end{tabular}
\caption{C-index on METABRIC for different SurvHL smoothing parameters and
time-grid sizes.}
\label{tab:survhl_grid}
\end{table}

\subsection{Comparison with external ensembles}
\label{sec:ensemble_ablation}

The LAS variant uses parameter-efficient ensembling with a shared embedding
module. To verify that its gains are not simply due to averaging many
independently trained models, we compare it with external ensembles of
DeepSurv, DeepHit, and LS(MLP), using 32 independently trained members where
applicable. Table~\ref{tab:ensemble_bayes} reports posterior probabilities of
superiority for the C-index, and Table~\ref{tab:ensemble_rope} reports the
corresponding ROPE probabilities. LS(MLP) already dominates DeepSurv with
high probability and remains competitive with DeepSurv Ensemble. The
independent LS(MLP) Ensemble is practically equivalent to LAS with high ROPE
probability, while LAS is trained as a single shared model and therefore
requires fewer computational resources.

\begin{table}
\centering
\small
\begin{tabular}{lrrrrrr}
\toprule
$P(A>B)$ & LS & LAS & LS Ens & DeepSurv & DS Ens & DH Ens \\
\midrule
LS & 0.000 & 0.225 & 0.162 & 0.978 & 0.564 & 0.926 \\
LAS & 0.011 & 0.000 & 0.049 & 0.910 & 0.502 & 0.941 \\
LS Ens & 0.161 & 0.112 & 0.000 & 0.967 & 0.835 & 0.972 \\
DeepSurv & 0.000 & 0.000 & 0.000 & 0.000 & 0.000 & 0.059 \\
DS Ens & 0.002 & 0.124 & 0.008 & 0.851 & 0.000 & 0.907 \\
DH Ens & 0.001 & 0.026 & 0.001 & 0.168 & 0.024 & 0.000 \\
\bottomrule
\end{tabular}
\caption{Bayesian C-index comparison between TabSurv variants and external
ensembles. LS denotes TabSurv\textsubscript{LS(MLP)}; DS and DH denote
DeepSurv and DeepHit.}
\label{tab:ensemble_bayes}
\end{table}

\begin{table}
\centering
\small
\begin{tabular}{lrrrrrr}
\toprule
$P(\mathrm{ROPE})$ & LS & LAS & LS Ens & DeepSurv & DS Ens & DH Ens \\
\midrule
LS & 1.000 & 0.764 & 0.674 & 0.022 & 0.434 & 0.073 \\
LAS & 0.760 & 1.000 & 0.840 & 0.090 & 0.370 & 0.033 \\
LS Ens & 0.678 & 0.838 & 1.000 & 0.033 & 0.157 & 0.026 \\
DeepSurv & 0.022 & 0.090 & 0.034 & 1.000 & 0.147 & 0.774 \\
DS Ens & 0.433 & 0.373 & 0.157 & 0.149 & 1.000 & 0.068 \\
DH Ens & 0.073 & 0.031 & 0.027 & 0.775 & 0.067 & 1.000 \\
\bottomrule
\end{tabular}
\caption{ROPE probabilities for the ensemble comparison in
Table~\ref{tab:ensemble_bayes}.}
\label{tab:ensemble_rope}
\end{table}

\section{Results of experiments on real data}\label{res_tables}

% \input{tables/METABRIC_table}
% \input{tables/SEER_table}
% \input{tables/SUPPORT_table}
% \input{tables/TCGA_GBM_table}
% \input{tables/rotterdam_table}
% \input{tables/flchain_table}
% \input{tables/gbsg2_table}
% \input{tables/whas500_table}
% \input{tables/lung_table}
% \input{tables/PBC_table}

% === Contents of METABRIC_table.tex ===
\begin{table}[H]
\centering
\footnotesize
\begin{tabular}{|c|c|c||c|c||c|c|}
\hline
\multirow{2}{*}{}& \multicolumn{2}{c||}{C-index}& \multicolumn{2}{c||}{IBS}& \multicolumn{2}{c|}{AUC}\\
\cline{2-7}& Mean ($\uparrow$) & Rank ($\downarrow$)& Mean ($\downarrow$) & Rank ($\downarrow$)& Mean ($\uparrow$) & Rank ($\downarrow$)
\\
\hline
$\text{TabSurv}_\text{LS(MLP)}$ & $\mathbf{0.815\pm0.003}$& $\mathbf{1.4\pm0.5}$ & $0.093\pm0.002$& $1.9\pm0.3$ & $\mathbf{0.836\pm0.007}$& $\mathbf{1.4\pm0.6}$ \\
\hline
$\text{TabSurv}_\text{LS(RealMLP)}$ & $0.713\pm0.004$& $7.4\pm1.1$ & $0.111\pm0.002$& $6.3\pm0.9$ & $0.755\pm0.005$& $6.8\pm0.8$ \\
\hline
$\text{TabSurv}_\text{LAS}$ & $0.812\pm0.003$& $1.9\pm0.7$ & $\mathbf{0.090\pm0.001}$& $\mathbf{1.1\pm0.3}$ & $0.832\pm0.003$& $1.8\pm0.6$ \\
\hline
$\text{TabSurv}_\text{WAS}$ & $0.805\pm0.004$& $3.3\pm0.5$ & $0.104\pm0.011$& $4.1\pm1.8$ & $0.793\pm0.008$& $4.0\pm0.7$ \\
\hline
$\text{TabSurv}_\text{WSA}$ & $0.788\pm0.006$& $4.8\pm0.4$ & $0.104\pm0.004$& $4.0\pm0.9$ & $0.779\pm0.009$& $5.0\pm0.9$ \\
\hline
DeepSurv & $0.709\pm0.017$& $7.5\pm1.1$ & $0.112\pm0.004$& $6.2\pm1.0$ & $0.733\pm0.016$& $8.2\pm0.8$ \\
\hline
DeepHit & $0.710\pm0.005$& $7.8\pm0.7$ & $0.171\pm0.001$& $8.9\pm0.2$ & $0.744\pm0.013$& $7.5\pm0.9$ \\
\hline
SurvTRACE & $0.695\pm0.072$& $7.3\pm1.3$ & $0.127\pm0.015$& $7.8\pm0.6$ & $0.722\pm0.076$& $6.8\pm2.1$ \\
\hline
RSF & $0.800\pm0.010$& $3.6\pm1.1$ & $0.106\pm0.002$& $4.7\pm0.8$ & $0.806\pm0.023$& $3.3\pm1.1$ \\
\hline
\end{tabular}
\caption{Comparison of model performance on the METABRIC dataset across C-index, Integrated Brier Score (IBS), and time-dependent AUC}
\label{tab:res_METABRIC}
\end{table}
% === Contents of SEER_table.tex ===
\begin{table}[H]
\centering
\footnotesize
\begin{tabular}{|c|c|c||c|c||c|c|}
\hline
\multirow{2}{*}{}& \multicolumn{2}{c||}{C-index}& \multicolumn{2}{c||}{IBS}& \multicolumn{2}{c|}{AUC}\\
\cline{2-7}& Mean ($\uparrow$) & Rank ($\downarrow$)& Mean ($\downarrow$) & Rank ($\downarrow$)& Mean ($\uparrow$) & Rank ($\downarrow$)
\\
\hline
$\text{TabSurv}_\text{LS(MLP)}$ & $0.724\pm0.007$& $6.7\pm2.0$ & $0.083\pm0.001$& $6.8\pm1.5$ & $0.738\pm0.012$& $6.1\pm2.8$ \\
\hline
$\text{TabSurv}_\text{LS(RealMLP)}$ & $\mathbf{0.734\pm0.002}$& $\mathbf{2.2\pm1.2}$ & $0.081\pm0.001$& $3.1\pm1.9$ & $\mathbf{0.752\pm0.003}$& $\mathbf{2.4\pm1.2}$ \\
\hline
$\text{TabSurv}_\text{LAS}$ & $0.733\pm0.004$& $3.4\pm2.2$ & $0.081\pm0.001$& $4.0\pm1.7$ & $0.751\pm0.004$& $3.0\pm1.8$ \\
\hline
$\text{TabSurv}_\text{WAS}$ & $0.729\pm0.003$& $5.3\pm1.7$ & $0.082\pm0.001$& $4.1\pm2.1$ & $0.745\pm0.005$& $5.0\pm1.8$ \\
\hline
$\text{TabSurv}_\text{WSA}$ & $0.731\pm0.003$& $4.2\pm2.4$ & $0.084\pm0.004$& $5.3\pm2.8$ & $0.748\pm0.003$& $4.1\pm1.6$ \\
\hline
DeepSurv & $0.729\pm0.005$& $4.9\pm2.1$ & $\mathbf{0.081\pm0.001}$& $\mathbf{2.4\pm1.4}$ & $0.748\pm0.005$& $3.8\pm1.8$ \\
\hline
DeepHit & $0.723\pm0.005$& $8.0\pm0.9$ & $0.209\pm0.010$& $9.0\pm0.0$ & $0.730\pm0.006$& $8.1\pm0.9$ \\
\hline
SurvTRACE & $0.725\pm0.008$& $5.8\pm2.9$ & $0.082\pm0.002$& $5.0\pm2.1$ & $0.739\pm0.013$& $5.5\pm2.9$ \\
\hline
RSF & $0.731\pm0.003$& $4.5\pm2.0$ & $0.082\pm0.000$& $5.3\pm1.2$ & $0.738\pm0.004$& $7.0\pm0.9$ \\
\hline
\end{tabular}
\caption{Comparison of model performance on the SEER dataset across C-index, Integrated Brier Score (IBS), and time-dependent AUC}
\label{tab:res_SEER}
\end{table}
% === Contents of SUPPORT_table.tex ===
\begin{table}[H]
\centering
\footnotesize
\begin{tabular}{|c|c|c||c|c||c|c|}
\hline
\multirow{2}{*}{}& \multicolumn{2}{c||}{C-index}& \multicolumn{2}{c||}{IBS}& \multicolumn{2}{c|}{AUC}\\
\cline{2-7}& Mean ($\uparrow$) & Rank ($\downarrow$)& Mean ($\downarrow$) & Rank ($\downarrow$)& Mean ($\uparrow$) & Rank ($\downarrow$)
\\
\hline
$\text{TabSurv}_\text{LS(MLP)}$ & $0.897\pm0.000$& $3.9\pm0.5$ & $0.096\pm0.002$& $2.7\pm1.2$ & $0.952\pm0.000$& $4.0\pm0.7$ \\
\hline
$\text{TabSurv}_\text{LS(RealMLP)}$ & $0.895\pm0.001$& $5.0\pm0.4$ & $0.099\pm0.002$& $4.5\pm1.4$ & $0.951\pm0.000$& $5.5\pm0.6$ \\
\hline
$\text{TabSurv}_\text{LAS}$ & $0.898\pm0.001$& $2.4\pm1.0$ & $0.100\pm0.005$& $4.6\pm2.4$ & $0.953\pm0.001$& $3.0\pm0.9$ \\
\hline
$\text{TabSurv}_\text{WAS}$ & $\mathbf{0.898\pm0.001}$& $\mathbf{1.6\pm0.9}$ & $0.097\pm0.005$& $3.0\pm2.5$ & $\mathbf{0.955\pm0.001}$& $\mathbf{1.4\pm0.7}$ \\
\hline
$\text{TabSurv}_\text{WSA}$ & $0.898\pm0.000$& $2.1\pm0.8$ & $\mathbf{0.096\pm0.002}$& $\mathbf{2.6\pm1.6}$ & $0.954\pm0.001$& $1.9\pm0.6$ \\
\hline
DeepSurv & $0.886\pm0.001$& $7.0\pm0.0$ & $0.103\pm0.001$& $6.5\pm0.5$ & $0.945\pm0.001$& $7.8\pm0.9$ \\
\hline
DeepHit & $0.879\pm0.002$& $8.0\pm0.0$ & $0.107\pm0.003$& $8.4\pm1.0$ & $0.945\pm0.001$& $8.3\pm0.7$ \\
\hline
SurvTRACE & $0.893\pm0.001$& $6.0\pm0.2$ & $0.100\pm0.004$& $4.8\pm2.0$ & $0.951\pm0.001$& $5.2\pm0.9$ \\
\hline
RSF & $0.858\pm0.001$& $9.0\pm0.0$ & $0.105\pm0.000$& $7.8\pm0.6$ & $0.945\pm0.001$& $7.8\pm0.7$ \\
\hline
\end{tabular}
\caption{Comparison of model performance on the SUPPORT dataset across C-index, Integrated Brier Score (IBS), and time-dependent AUC}
\label{tab:res_SUPPORT}
\end{table}
% === Contents of TCGA_GBM_table.tex ===
\begin{table}[H]
\centering
\footnotesize
\begin{tabular}{|c|c|c||c|c||c|c|}
\hline
\multirow{2}{*}{}& \multicolumn{2}{c||}{C-index}& \multicolumn{2}{c||}{IBS}& \multicolumn{2}{c|}{AUC}\\
\cline{2-7}& Mean ($\uparrow$) & Rank ($\downarrow$)& Mean ($\downarrow$) & Rank ($\downarrow$)& Mean ($\uparrow$) & Rank ($\downarrow$)
\\
\hline
$\text{TabSurv}_\text{LS(MLP)}$ & $0.863\pm0.003$& $3.1\pm1.2$ & $\mathbf{0.062\pm0.001}$& $\mathbf{1.9\pm0.7}$ & $0.918\pm0.003$& $4.4\pm1.5$ \\
\hline
$\text{TabSurv}_\text{LS(RealMLP)}$ & $0.853\pm0.003$& $6.0\pm1.0$ & $0.063\pm0.001$& $2.2\pm1.0$ & $0.913\pm0.003$& $6.5\pm1.3$ \\
\hline
$\text{TabSurv}_\text{LAS}$ & $0.861\pm0.007$& $3.8\pm1.9$ & $0.066\pm0.006$& $3.5\pm2.2$ & $0.915\pm0.008$& $5.7\pm2.4$ \\
\hline
$\text{TabSurv}_\text{WAS}$ & $0.861\pm0.003$& $3.7\pm1.1$ & $0.065\pm0.005$& $3.2\pm1.4$ & $0.917\pm0.003$& $4.7\pm1.5$ \\
\hline
$\text{TabSurv}_\text{WSA}$ & $0.862\pm0.003$& $3.5\pm1.0$ & $0.074\pm0.009$& $6.0\pm1.8$ & $0.918\pm0.005$& $4.5\pm1.6$ \\
\hline
DeepSurv & $0.841\pm0.008$& $8.0\pm1.0$ & $0.072\pm0.002$& $5.9\pm1.0$ & $0.892\pm0.009$& $8.8\pm0.5$ \\
\hline
DeepHit & $\mathbf{0.868\pm0.003}$& $\mathbf{1.3\pm0.6}$ & $0.084\pm0.001$& $8.7\pm0.7$ & $0.923\pm0.003$& $2.0\pm1.2$ \\
\hline
SurvTRACE & $0.837\pm0.011$& $8.3\pm0.9$ & $0.077\pm0.010$& $7.0\pm1.3$ & $0.910\pm0.009$& $6.7\pm2.2$ \\
\hline
RSF & $0.847\pm0.004$& $7.2\pm0.8$ & $0.073\pm0.001$& $6.5\pm1.0$ & $\mathbf{0.924\pm0.002}$& $\mathbf{1.9\pm0.8}$ \\
\hline
\end{tabular}
\caption{Comparison of model performance on the TCGA-GBM dataset across C-index, Integrated Brier Score (IBS), and time-dependent AUC}
\label{tab:res_TCGA_GBM}
\end{table}
% === Contents of rotterdam_table.tex ===
\begin{table}[H]
\centering
\footnotesize
\begin{tabular}{|c|c|c||c|c||c|c|}
\hline
\multirow{2}{*}{}& \multicolumn{2}{c||}{C-index}& \multicolumn{2}{c||}{IBS}& \multicolumn{2}{c|}{AUC}\\
\cline{2-7}& Mean ($\uparrow$) & Rank ($\downarrow$)& Mean ($\downarrow$) & Rank ($\downarrow$)& Mean ($\uparrow$) & Rank ($\downarrow$)
\\
\hline
$\text{TabSurv}_\text{LS(MLP)}$ & $0.723\pm0.002$& $2.7\pm1.5$ & $0.145\pm0.001$& $2.5\pm0.7$ & $0.775\pm0.003$& $3.5\pm1.3$ \\
\hline
$\text{TabSurv}_\text{LS(RealMLP)}$ & $0.721\pm0.002$& $3.9\pm1.4$ & $0.148\pm0.002$& $4.5\pm1.6$ & $0.777\pm0.003$& $2.6\pm1.2$ \\
\hline
$\text{TabSurv}_\text{LAS}$ & $0.722\pm0.002$& $3.4\pm1.3$ & $\mathbf{0.142\pm0.000}$& $\mathbf{1.1\pm0.2}$ & $\mathbf{0.780\pm0.003}$& $\mathbf{1.4\pm0.9}$ \\
\hline
$\text{TabSurv}_\text{WAS}$ & $\mathbf{0.723\pm0.002}$& $\mathbf{2.5\pm1.3}$ & $0.151\pm0.008$& $4.8\pm2.0$ & $0.771\pm0.005$& $4.7\pm1.6$ \\
\hline
$\text{TabSurv}_\text{WSA}$ & $0.723\pm0.002$& $2.7\pm1.4$ & $0.155\pm0.012$& $6.0\pm2.2$ & $0.770\pm0.004$& $5.3\pm1.1$ \\
\hline
DeepSurv & $0.708\pm0.003$& $7.3\pm0.5$ & $0.151\pm0.001$& $6.7\pm0.9$ & $0.754\pm0.003$& $7.8\pm0.8$ \\
\hline
DeepHit & $0.706\pm0.003$& $7.7\pm0.6$ & $0.209\pm0.007$& $8.8\pm0.4$ & $0.754\pm0.004$& $8.0\pm0.7$ \\
\hline
SurvTRACE & $0.675\pm0.015$& $9.0\pm0.0$ & $0.191\pm0.104$& $5.5\pm2.4$ & $0.750\pm0.014$& $7.8\pm1.8$ \\
\hline
RSF & $0.717\pm0.002$& $5.8\pm0.7$ & $0.148\pm0.001$& $5.1\pm1.0$ & $0.773\pm0.003$& $4.0\pm1.0$ \\
\hline
\end{tabular}
\caption{Comparison of model performance on the Rotterdam dataset across C-index, Integrated Brier Score (IBS), and time-dependent AUC}
\label{tab:res_rotterdam}
\end{table}
% === Contents of flchain_table.tex ===
\begin{table}[H]
\centering
\footnotesize
\begin{tabular}{|c|c|c||c|c||c|c|}
\hline
\multirow{2}{*}{}& \multicolumn{2}{c||}{C-index}& \multicolumn{2}{c||}{IBS}& \multicolumn{2}{c|}{AUC}\\
\cline{2-7}& Mean ($\uparrow$) & Rank ($\downarrow$)& Mean ($\downarrow$) & Rank ($\downarrow$)& Mean ($\uparrow$) & Rank ($\downarrow$)
\\
\hline
$\text{TabSurv}_\text{LS(MLP)}$ & $0.936\pm0.001$& $5.0\pm1.2$ & $0.047\pm0.005$& $3.5\pm1.4$ & $0.955\pm0.001$& $4.2\pm1.6$ \\
\hline
$\text{TabSurv}_\text{LS(RealMLP)}$ & $0.935\pm0.001$& $6.4\pm1.0$ & $0.047\pm0.001$& $5.5\pm0.9$ & $0.951\pm0.002$& $7.8\pm1.2$ \\
\hline
$\text{TabSurv}_\text{LAS}$ & $0.937\pm0.001$& $4.3\pm1.1$ & $0.046\pm0.000$& $3.6\pm1.5$ & $0.955\pm0.001$& $4.5\pm1.4$ \\
\hline
$\text{TabSurv}_\text{WAS}$ & $\mathbf{0.938\pm0.000}$& $\mathbf{1.6\pm0.7}$ & $0.046\pm0.002$& $3.8\pm1.9$ & $\mathbf{0.957\pm0.000}$& $\mathbf{1.6\pm0.5}$ \\
\hline
$\text{TabSurv}_\text{WSA}$ & $0.938\pm0.000$& $2.1\pm0.7$ & $0.047\pm0.002$& $4.7\pm2.1$ & $0.950\pm0.013$& $3.0\pm3.1$ \\
\hline
DeepSurv & $0.938\pm0.000$& $2.5\pm1.0$ & $\mathbf{0.044\pm0.000}$& $\mathbf{1.1\pm0.4}$ & $0.955\pm0.001$& $3.6\pm0.9$ \\
\hline
DeepHit & $0.933\pm0.001$& $8.4\pm0.7$ & $0.055\pm0.001$& $8.8\pm0.7$ & $0.952\pm0.001$& $7.4\pm1.5$ \\
\hline
SurvTRACE & $0.930\pm0.014$& $7.5\pm1.5$ & $0.055\pm0.021$& $6.8\pm1.6$ & $0.934\pm0.081$& $7.1\pm1.2$ \\
\hline
RSF & $0.934\pm0.001$& $7.0\pm1.2$ & $0.048\pm0.000$& $7.1\pm0.8$ & $0.953\pm0.001$& $5.8\pm1.5$ \\
\hline
\end{tabular}
\caption{Comparison of model performance on the FLC dataset across C-index, Integrated Brier Score (IBS), and time-dependent AUC}
\label{tab:res_flchain}
\end{table}
% === Contents of gbsg2_table.tex ===
\begin{table}[H]
\centering
\footnotesize
\begin{tabular}{|c|c|c||c|c||c|c|}
\hline
\multirow{2}{*}{}& \multicolumn{2}{c||}{C-index}& \multicolumn{2}{c||}{IBS}& \multicolumn{2}{c|}{AUC}\\
\cline{2-7}& Mean ($\uparrow$) & Rank ($\downarrow$)& Mean ($\downarrow$) & Rank ($\downarrow$)& Mean ($\uparrow$) & Rank ($\downarrow$)
\\
\hline
$\text{TabSurv}_\text{LS(MLP)}$ & $\mathbf{0.719\pm0.007}$& $\mathbf{2.2\pm1.4}$ & $0.152\pm0.010$& $\mathbf{2.2\pm1.5}$ & $0.756\pm0.027$& $5.6\pm1.1$ \\
\hline
$\text{TabSurv}_\text{LS(RealMLP)}$ & $0.702\pm0.005$& $6.6\pm1.2$ & $0.170\pm0.009$& $6.5\pm1.0$ & $0.762\pm0.015$& $5.3\pm1.1$ \\
\hline
$\text{TabSurv}_\text{LAS}$ & $0.714\pm0.007$& $3.9\pm1.8$ & $0.159\pm0.007$& $4.5\pm2.0$ & $0.765\pm0.007$& $5.0\pm1.2$ \\
\hline
$\text{TabSurv}_\text{WAS}$ & $0.716\pm0.004$& $3.2\pm1.4$ & $0.156\pm0.005$& $4.2\pm1.1$ & $0.788\pm0.004$& $1.9\pm0.7$ \\
\hline
$\text{TabSurv}_\text{WSA}$ & $0.714\pm0.006$& $3.8\pm1.6$ & $0.169\pm0.043$& $5.4\pm1.1$ & $0.784\pm0.006$& $2.4\pm0.7$ \\
\hline
DeepSurv & $0.700\pm0.014$& $6.3\pm2.1$ & $\mathbf{0.152\pm0.003}$& $2.6\pm1.5$ & $0.755\pm0.019$& $6.0\pm1.6$ \\
\hline
DeepHit & $0.669\pm0.026$& $8.7\pm0.6$ & $0.202\pm0.002$& $8.0\pm0.2$ & $0.689\pm0.033$& $7.9\pm0.3$ \\
\hline
SurvTRACE & $0.685\pm0.037$& $6.9\pm2.0$ & $0.607\pm0.000$& $9.0\pm0.0$ & $0.548\pm0.007$& $9.0\pm0.0$ \\
\hline
RSF & $0.715\pm0.005$& $3.3\pm1.6$ & $0.152\pm0.001$& $2.6\pm0.9$ & $\mathbf{0.788\pm0.009}$& $\mathbf{1.9\pm1.0}$ \\
\hline
\end{tabular}
\caption{Comparison of model performance on the GBSG2 dataset across C-index, Integrated Brier Score (IBS), and time-dependent AUC}
\label{tab:res_gbsg2}
\end{table}
% === Contents of whas500_table.tex ===
\begin{table}[H]
\centering
\footnotesize
\begin{tabular}{|c|c|c||c|c||c|c|}
\hline
\multirow{2}{*}{}& \multicolumn{2}{c||}{C-index}& \multicolumn{2}{c||}{IBS}& \multicolumn{2}{c|}{AUC}\\
\cline{2-7}& Mean ($\uparrow$) & Rank ($\downarrow$)& Mean ($\downarrow$) & Rank ($\downarrow$)& Mean ($\uparrow$) & Rank ($\downarrow$)
\\
\hline
$\text{TabSurv}_\text{LS(MLP)}$ & $\mathbf{0.801\pm0.026}$& $\mathbf{2.0\pm1.8}$ & $0.163\pm0.018$& $3.8\pm2.0$ & $0.826\pm0.044$& $3.4\pm2.4$ \\
\hline
$\text{TabSurv}_\text{LS(RealMLP)}$ & $0.763\pm0.017$& $8.0\pm1.1$ & $0.200\pm0.016$& $7.4\pm0.8$ & $0.803\pm0.020$& $7.0\pm1.8$ \\
\hline
$\text{TabSurv}_\text{LAS}$ & $0.782\pm0.006$& $6.0\pm1.1$ & $0.163\pm0.008$& $4.5\pm1.7$ & $0.828\pm0.008$& $4.3\pm1.5$ \\
\hline
$\text{TabSurv}_\text{WAS}$ & $0.801\pm0.009$& $2.7\pm1.7$ & $\mathbf{0.153\pm0.011}$& $\mathbf{2.0\pm2.1}$ & $0.826\pm0.029$& $3.8\pm2.2$ \\
\hline
$\text{TabSurv}_\text{WSA}$ & $0.795\pm0.013$& $4.1\pm2.5$ & $0.198\pm0.074$& $4.3\pm2.6$ & $0.809\pm0.035$& $5.9\pm2.6$ \\
\hline
DeepSurv & $0.783\pm0.018$& $5.8\pm2.1$ & $0.160\pm0.010$& $3.5\pm1.6$ & $0.811\pm0.020$& $6.2\pm2.3$ \\
\hline
DeepHit & $0.780\pm0.013$& $6.2\pm1.9$ & $0.194\pm0.025$& $6.5\pm1.7$ & $0.820\pm0.018$& $5.7\pm2.2$ \\
\hline
SurvTRACE & $0.777\pm0.024$& $6.2\pm2.6$ & $0.350\pm0.153$& $8.2\pm1.6$ & $0.801\pm0.063$& $5.8\pm2.7$ \\
\hline
RSF & $0.793\pm0.006$& $4.1\pm1.2$ & $0.162\pm0.003$& $4.8\pm1.0$ & $\mathbf{0.839\pm0.008}$& $\mathbf{2.8\pm1.7}$ \\
\hline
\end{tabular}
\caption{Comparison of model performance on the WHAS500 dataset across C-index, Integrated Brier Score (IBS), and time-dependent AUC}
\label{tab:res_whas500}
\end{table}
% === Contents of lung_table.tex ===
\begin{table}[H]
\centering
\footnotesize
\begin{tabular}{|c|c|c||c|c||c|c|}
\hline
\multirow{2}{*}{}& \multicolumn{2}{c||}{C-index}& \multicolumn{2}{c||}{IBS}& \multicolumn{2}{c|}{AUC}\\
\cline{2-7}& Mean ($\uparrow$) & Rank ($\downarrow$)& Mean ($\downarrow$) & Rank ($\downarrow$)& Mean ($\uparrow$) & Rank ($\downarrow$)
\\
\hline
$\text{TabSurv}_\text{LS(MLP)}$ & $0.649\pm0.012$& $2.3\pm0.8$ & $0.166\pm0.003$& $2.9\pm1.5$ & $0.646\pm0.014$& $3.0\pm1.0$ \\
\hline
$\text{TabSurv}_\text{LS(RealMLP)}$ & $0.588\pm0.045$& $5.4\pm1.6$ & $0.167\pm0.002$& $3.9\pm1.2$ & $0.604\pm0.034$& $4.8\pm1.2$ \\
\hline
$\text{TabSurv}_\text{LAS}$ & $0.621\pm0.037$& $3.9\pm1.8$ & $0.169\pm0.006$& $4.3\pm1.3$ & $0.630\pm0.022$& $3.8\pm1.3$ \\
\hline
$\text{TabSurv}_\text{WAS}$ & $0.571\pm0.032$& $6.2\pm1.3$ & $0.295\pm0.077$& $7.8\pm0.7$ & $0.528\pm0.039$& $7.1\pm1.2$ \\
\hline
$\text{TabSurv}_\text{WSA}$ & $0.517\pm0.042$& $7.8\pm1.5$ & $0.392\pm0.081$& $8.8\pm0.4$ & $0.486\pm0.029$& $8.2\pm0.9$ \\
\hline
DeepSurv & $0.624\pm0.024$& $4.0\pm1.4$ & $0.170\pm0.004$& $5.0\pm1.7$ & $0.648\pm0.032$& $2.8\pm1.5$ \\
\hline
DeepHit & $0.517\pm0.061$& $7.5\pm1.7$ & $\mathbf{0.165\pm0.001}$& $\mathbf{2.4\pm1.1}$ & $0.519\pm0.045$& $7.4\pm1.2$ \\
\hline
SurvTRACE & $0.549\pm0.082$& $6.5\pm2.2$ & $0.210\pm0.082$& $6.2\pm2.3$ & $0.556\pm0.081$& $6.2\pm2.6$ \\
\hline
RSF & $\mathbf{0.668\pm0.018}$& $\mathbf{1.5\pm1.2}$ & $0.166\pm0.004$& $3.7\pm2.3$ & $\mathbf{0.670\pm0.023}$& $\mathbf{1.8\pm1.0}$ \\
\hline
\end{tabular}
\caption{Comparison of model performance on the Lung dataset across C-index, Integrated Brier Score (IBS), and time-dependent AUC}
\label{tab:res_lung}
\end{table}
% === Contents of PBC_table.tex ===
\begin{table}[H]
\centering
\footnotesize
\begin{tabular}{|c|c|c||c|c||c|c|}
\hline
\multirow{2}{*}{}& \multicolumn{2}{c||}{C-index}& \multicolumn{2}{c||}{IBS}& \multicolumn{2}{c|}{AUC}\\
\cline{2-7}& Mean ($\uparrow$) & Rank ($\downarrow$)& Mean ($\downarrow$) & Rank ($\downarrow$)& Mean ($\uparrow$) & Rank ($\downarrow$)
\\
\hline
$\text{TabSurv}_\text{LS(MLP)}$ & $0.981\pm0.013$& $5.4\pm1.0$ & $0.046\pm0.050$& $4.5\pm1.2$ & $0.985\pm0.013$& $5.5\pm1.1$ \\
\hline
$\text{TabSurv}_\text{LS(RealMLP)}$ & $0.969\pm0.017$& $7.0\pm0.5$ & $0.057\pm0.046$& $6.0\pm0.7$ & $0.955\pm0.011$& $8.6\pm0.5$ \\
\hline
$\text{TabSurv}_\text{LAS}$ & $0.993\pm0.002$& $3.4\pm0.5$ & $0.033\pm0.004$& $4.0\pm0.9$ & $0.992\pm0.007$& $3.9\pm1.1$ \\
\hline
$\text{TabSurv}_\text{WAS}$ & $\mathbf{0.997\pm0.001}$& $1.6\pm0.6$ & $\mathbf{0.023\pm0.006}$& $\mathbf{1.7\pm1.1}$ & $\mathbf{0.999\pm0.001}$& $1.8\pm0.8$ \\
\hline
$\text{TabSurv}_\text{WSA}$ & $0.997\pm0.001$& $\mathbf{1.5\pm0.5}$ & $0.029\pm0.009$& $2.7\pm1.5$ & $0.999\pm0.001$& $\mathbf{1.6\pm0.9}$ \\
\hline
DeepSurv & $0.983\pm0.002$& $5.3\pm0.6$ & $0.028\pm0.002$& $2.4\pm0.8$ & $0.991\pm0.002$& $4.9\pm0.9$ \\
\hline
DeepHit & $0.949\pm0.029$& $8.5\pm0.5$ & $0.232\pm0.011$& $8.6\pm0.5$ & $0.952\pm0.042$& $7.7\pm0.9$ \\
\hline
SurvTRACE & $0.977\pm0.054$& $4.2\pm1.7$ & $0.219\pm0.229$& $8.0\pm0.8$ & $0.993\pm0.010$& $3.5\pm1.5$ \\
\hline
RSF & $0.964\pm0.004$& $8.1\pm0.8$ & $0.069\pm0.003$& $7.1\pm0.4$ & $0.967\pm0.007$& $7.5\pm0.7$ \\
\hline
\end{tabular}
\caption{Comparison of model performance on the PBC dataset across C-index, Integrated Brier Score (IBS), and time-dependent AUC}
\label{tab:res_PBC}
\end{table}

\section{Simulation Experiments}
\label{sec:simulation}

As demonstrated in the experiments on real-world datasets, 
TabSurv\textsubscript{WAS} achieves the best performance in terms of C-index and time-dependent AUC, 
while remaining competitive with respect to IBS. 
However, this behavior may be influenced by the specific characteristics of the considered real datasets. 
To further investigate the limitations of the proposed methods, we conduct an additional simulation study 
designed to illustrate scenarios in which TabSurv\textsubscript{WAS} is clearly less effective than 
TabSurv\textsubscript{LAS} with respect to the remaining metrics.

Synthetic event times were generated using the covariates of the Rotterdam dataset. 
To induce a bimodal time-to-event distribution, each feature vector was randomly assigned to one of two latent clusters 
using a Bernoulli distribution, with approximately equal cluster sizes. 
Conditioned on the cluster assignment, event times were sampled from a covariate-dependent Weibull distribution 
following the procedure described in \cite{bender2005generating}, with parameters defined as
\begin{equation}
\begin{split}
    \lambda_0 = 10^{-5},\quad&\lambda_1=10^{-10},\\
    k_0=4,\quad&k_1=6,\\
    \beta_0\sim \mathrm{U}[0,1],\quad&\beta_1\sim \mathrm{U}[-1, 1].
\end{split}
\end{equation}

The random cluster assignment induces a bimodal event-time distribution for each covariate vector $\mathbf{x}$. 
Censoring indicators were generated independently using a Bernoulli distribution with a predefined censoring rate. 
The data were split into training, validation, and test sets using the same protocol as in the original Rotterdam experiments.

The goal of this experiment is to assess how well different models approximate the underlying bimodal event-time distribution. 
To this end, we employ the Kolmogorov--Smirnov (KS) test to statistically quantify the similarity between 
the empirical distribution of test event times and the distribution obtained by averaging the predicted discrete 
probability densities across the test set. 
The KS statistic is computed exclusively on the test subset. 
Specifically, each model predicts a discrete event-time distribution for each test instance on a time grid defined by 
the event times observed in the training set. 
These predicted distributions are then averaged across all test instances to obtain a population-level prior, 
which is subsequently compared to the empirical distribution of observed test event times.

\begin{figure}[t]
    \centering
    \includegraphics[width=\linewidth]{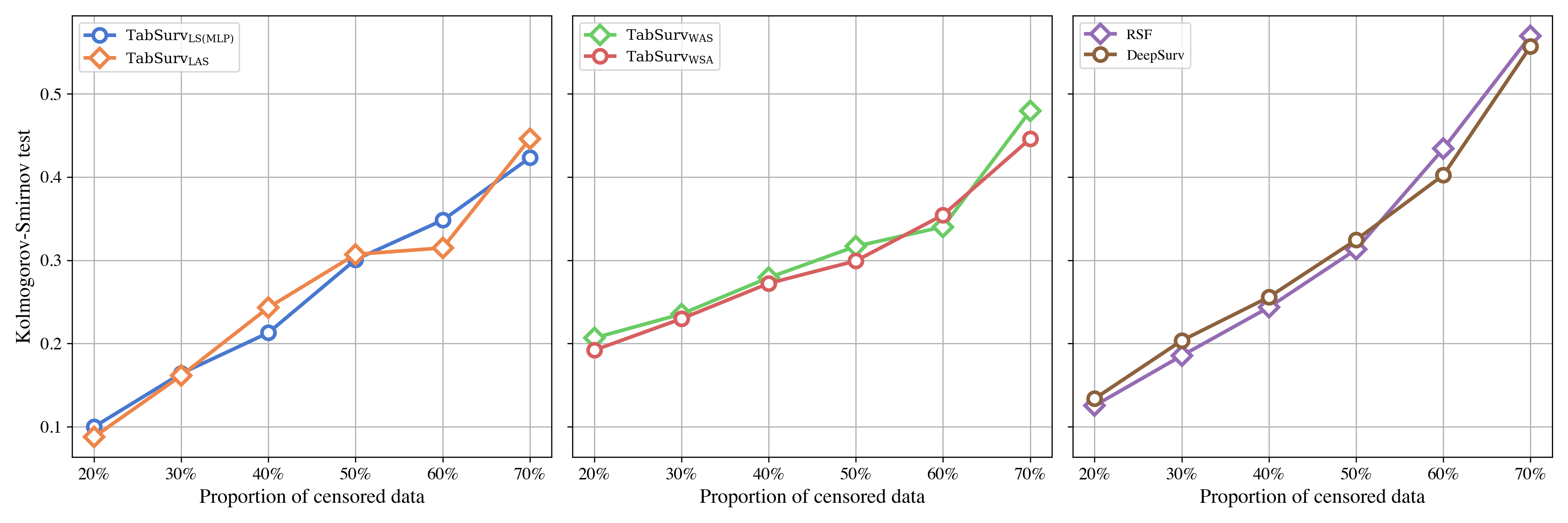}
    \caption{Results of the simulation study evaluated using the Kolmogorov--Smirnov statistic.}
    \label{fig:sim_figure}
\end{figure}

The censoring proportion is varied, and each point in Figure~\ref{fig:sim_figure} corresponds to the average over 
100 independent experimental runs. 
Standard deviations are also reported to illustrate the variability of the results.

\begin{figure}[t]
    \centering
    \includegraphics[width=\linewidth]{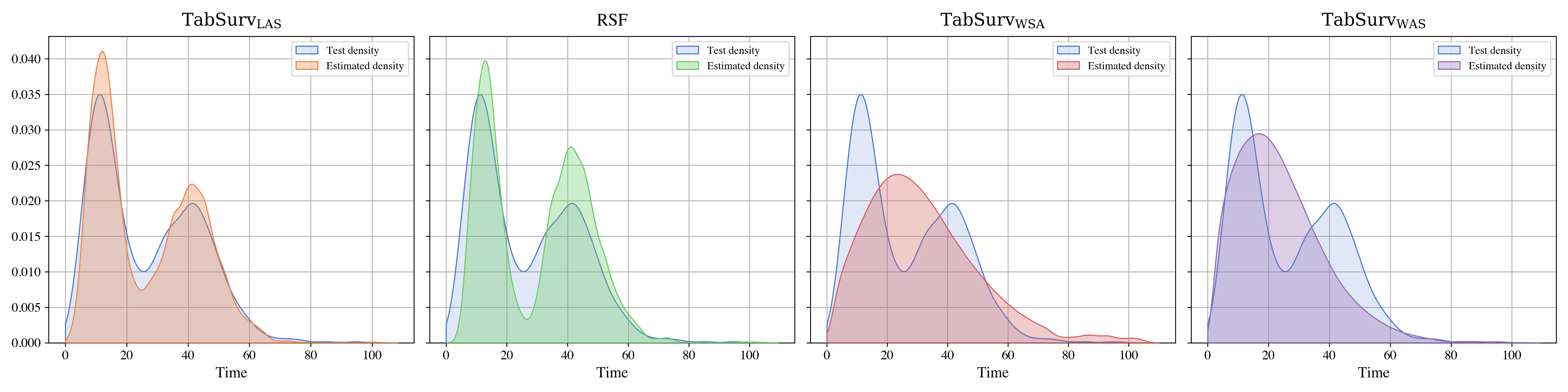}
    \caption{Predicted event-time densities on the simulated dataset for TabSurv\textsubscript{LAS}, RSF, and TabSurv\textsubscript{WAS}.}
    \label{fig:sim_density}
\end{figure}

Figure~\ref{fig:sim_density} visualizes the predicted event-time densities for a single experimental run with 
20\% censoring, using the SurvHL-TabM, RSF, and SurvHL-WPRM models. 
Kernel density estimation is used for visualization, as the original discrete distributions consist of approximately 
1{,}000 time intervals, making histogram-based representations difficult to interpret. 
This figure provides qualitative insight into how different models capture the shape of the simulated event-time distribution.

For clarity of presentation, only a subset of models is included in this experiment, namely 
TabSurv\textsubscript{LS(MLP)}, TabSurv\textsubscript{LAS}, RSF, TabSurv\textsubscript{WSA}, TabSurv\textsubscript{WAS}, and DeepSurv. 
These models were selected based on their performance in the real-data experiments.

Figure~\ref{fig:sim_figure} shows that the TabSurv\textsubscript{LS} and TabSurv\textsubscript{LAS} models, which are not constrained by a Weibull 
distributional assumption, consistently achieve superior performance across all censoring proportions. 
The event-time distributions predicted by these models are the closest to the empirical distribution observed in the test data.

Figure~\ref{fig:sim_density} highlights substantial differences in the shapes of the distributions predicted by the 
considered models. 
Among them, TabSurv\textsubscript{LAS} most closely matches the ground-truth bimodal distribution, providing the most accurate 
approximation in this setting.

%You can have as much text here as you want. The main body must be at most $8$
%pages long. For the final version, one more page can be added. If you want,
%you can use an appendix like this one.
%
%The $\mathtt{\backslash onecolumn}$ command above can be kept in place if you
%prefer a one-column appendix, or can be removed if you prefer a two-column
%appendix. Apart from this possible change, the style (font size, spacing,
%margins, page numbering, etc.) should be kept the same as the main body.

\end{document}